%% file: main_with_supp.tex
\documentclass[runningheads]{llncs}

\usepackage{eccv}

\usepackage{eccvabbrv}

\usepackage{graphicx}
\usepackage{booktabs}
\usepackage{multirow}
\usepackage{amssymb}
\usepackage{color, colortbl}

\usepackage[accsupp]{axessibility}  

\usepackage[breaklinks,colorlinks,citecolor=eccvblue]{hyperref}

\usepackage{orcidlink}
\usepackage{wrapfig}

\usepackage{pifont}
\newcommand{\cmark}{\ding{51}}%
\newcommand{\xmark}{\ding{55}}%

\newcommand{\myheading}[1]{\noindent \textbf{#1}}

\let\titleold\title
\renewcommand{\title}[1]{\titleold{#1}\newcommand{\thetitle}{#1}}
\def\maketitlesupplementary
   {
   \newpage
       \begin{center}
        \Large
        \textbf{\thetitle\\-- Supplementary Materials --}\\
        \vspace{1.0em}
       \end{center}
   }

\begin{document}


\title{SwiftBrush v2: Make Your One-step Diffusion Model Better Than Its Teacher} 

\titlerunning{SwiftBrush v2}

\author{Trung Dao\inst{1}\orcidlink{0009-0002-5217-4558} \quad
Thuan Hoang Nguyen\inst{1,*}\orcidlink{0009-0009-8012-5463} \quad
Thanh Le\inst{1,*}\orcidlink{0009-0008-7015-4212}
\quad Duc Vu\inst{1,*}\orcidlink{0009-0005-9639-0529} \\
\quad Khoi Nguyen\inst{1}\orcidlink{0000-0002-9259-420X}
\quad Cuong Pham\inst{1,2}\orcidlink{0000-0003-0973-0889}
\quad Anh Tran\inst{1}\orcidlink{0000-0002-3120-4036}}

\authorrunning{Dao et al.}

\institute{$^1$VinAI Research \quad $^2$Posts \& Telecommunications Inst. of Tech.}


\maketitle

\begin{abstract} 
In this paper, we aim to enhance the performance of SwiftBrush, a prominent one-step text-to-image diffusion model, to be competitive with its multi-step Stable Diffusion counterpart. Initially, we explore the quality-diversity trade-off between SwiftBrush and SD Turbo: the former excels in image diversity, while the latter excels in image quality. This observation motivates our proposed modifications in the training methodology, including better weight initialization and efficient LoRA training. Moreover, our introduction of a novel clamped CLIP loss enhances image-text alignment and results in improved image quality. Remarkably, by combining the weights of models trained with efficient LoRA and full training, we achieve a new state-of-the-art one-step diffusion model, achieving an FID of 8.14 and surpassing all GAN-based and multi-step Stable Diffusion models. The project page is available at: \url{https://swiftbrushv2.github.io/}

\makeatletter\def\Hy@Warning#1{}\makeatother
\def\thefootnote{*}\footnotetext{Equal Contribution.}

\end{abstract}

\section{Introduction}
\label{sec:intro}
Text-to-image generation has experienced tremendous growth in recent years, allowing users to create high-quality images from simple descriptions. State-of-the-art models \cite{LDM,DALLE2,DALLE3,Midjourney} could surpass humans in art competition \cite{Roose2022Sep} or produce synthetic images nearly indistinguishable from real ones \cite{Check2023Mar}. Among popular text-to-image networks, Stable Diffusion (SD) models \cite{LDM,SD} are the most widely used due to their open-source accessibility. However, most SD models are designed as multi-step diffusion models, which require multiple forwarding steps to produce an output image. Such a slow and computationally expensive mechanism hinders the use of these models in real-time or on-device applications.

Recently, many works have tried to reduce the denoising steps required in text-to-image diffusion models. Notably, few recent studies have successfully developed \textbf{one-step diffusion models}, thus significantly speed up the image generation. While early attempts \cite{LCM,BOOT} produce blurry and malformed photos, subsequent methods produce sharp and high-quality outputs. These methods mainly distill knowledge from a pre-trained multi-step SD model (referred to as the \textit{teacher model}) to a one-step \textit{student model}. InstaFlow \cite{InstaFlow} employs Rectified Flows \cite{liu2023flow} in a multi-stage and computation-expensive training procedure. DMD \cite{DMD} combines a reconstruction and a distribution matching loss as the training objectives, requiring massive pre-generated images from the teacher. SD Turbo \cite{SDTurbo} incorporates adversarial training alongside a score distillation loss, achieving photorealistic generation. However, it heavily relies on a large-scale image-text pair training dataset and, as later discussed, has a poor diversity. 
SwiftBrush \cite{SwiftBrush} utilizes Variational Score Distillation (VSD) to transfer knowledge from the teacher network to the one-step student through a LoRA \cite{hu2021lora} intermediate teacher model. Notably, training SwiftBrush is simple, fast, and image-free, making it an intriguing method. 

\begin{figure}[t]
    \centering
    \includegraphics[width=\textwidth]{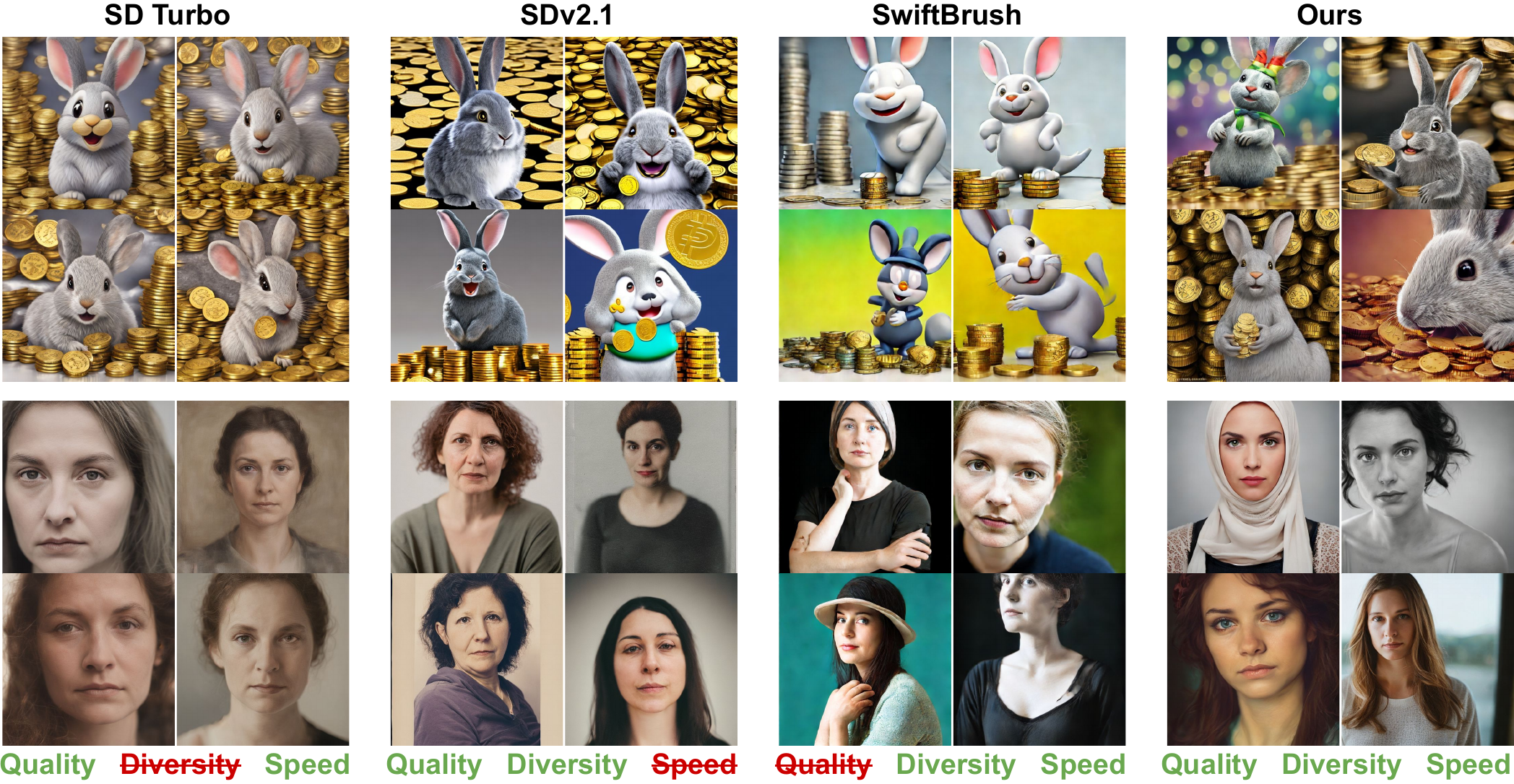}
    \caption{Our one-step diffusion model achieves an impressive FID of 8.14, generating high-quality and diverse results with a single UNet forwarding. The example images generated from the ``A laughing cute grey rabbit with white stripe on the head, piles of gold coins in background, colorful, Disney Picture render, photorealistic'' (first two rows) and ``Portrait of a woman looking at the camera'' (last two rows) prompts demonstrate our model's ability to create fast, visually appealing, and varied outputs.}
    \label{fig:teaser}
\end{figure}

Despite these promising achievements, one-step text-to-image diffusion models still fall short of multi-step models in terms of the FID metric. On the standard COCO 2014 benchmark \cite{lin2014microsoft}, SDv2.1 can achieve the lowest FID-30K score of 9.64 with classifier-free guidance (\textit{cfg}) scale of 2, while the best-reported score from the one-step models of equivalent parameters scale is 11.49 \cite{DMD}.
The gap is expected since directly predicting a clean image from noise in a single step is much more challenging than via a multi-step scheme. Hence, one may believe that one-step text-to-image models could only approach or reach a similar performance as the teacher model but never exceeding it.

In this paper, we challenge this belief by seeking a one-step model that can surpass its multi-step teacher model quantitatively and qualitatively. 
%
Our solution drew inspiration from SwiftBrush, with its image-free training, enabling an effective, scalable, and flexible distillation. We examine its current state and compare it with SD Turbo. The comparison in ~\cref{sec:analysis} highlights a quality-diversity trade-off: SwiftBrush offers more diverse outputs due to its image-free and dynamic training, while SD Turbo yields high-quality but mode-collapse-like outputs due to its adversarial training. This insight drives us to initialize SwiftBrush training with SD Turbo, enhancing one-step student models significantly. Moreover, our extra clamped CLIP loss combined with SwiftBrush's flexible training mechanism empowers the student model to surpass the teacher. 
Lastly, we train the student on a larger text prompt dataset for better knowledge transfer between teacher and student models.

Given limited resources and the goal of offering effective and affordable model training solutions, we restrict the training
on A100 40GB GPUs with affordable GPU hours.
Such a restricted condition prevents us from employing efficiently the clamped CLIP loss and fully finetuning the student model. Hence, we propose two training schemes, one supporting full student model training without the extra loss and one employing LoRA-based model training associated with the mentioned auxiliary loss. Both training schemes produce high-quality output models that surpass all previous one-step diffusion-based approaches in most metrics. Especially when merging these two models using a simple weight linear interpolation, we obtain an one-step model with FID-30K of \textbf{8.77} on the COCO 2014 benchmark. Such a student model is the first to surpass its multi-step teacher model,
breaking the common belief. It even exceeds all GAN-based text-to-image approaches \cite{sauer2023stylegan,kang2023scaling} while also offering near real-time speed. With an extra regularization \cite{DMD} on minimal real data, our merged model gets enhanced further to achieve the \textbf{FID score of 8.14}, setting a new standard for efficient and high-quality text-to-image models. 

In summary, our contributions include (1) an analysis of representative existing diffusion-based text-to-image models to reveal the quality-diversity trade-off, (2) a simple but effective integration of SwiftBrush and SD Turbo to combine the advantages of both, (3) an extra clamped CLIP loss proposed to boost the image-text alignment of the student network and surpass the teacher model, (4) two resource-efficient training strategies to utilize the mentioned proposals, and (5) a fused one-step student model that is superior to its multi-step text-to-image teacher in all metrics and sets a new standard in this field.

\section{Related Work}
\subsection{Text-to-Image Generation}
Text-to-image generation involves synthesizing high-quality images based on input text prompts. This task has evolved over decades, transitioning from constrained domains like CUB
\cite{CUB} and COCO
\cite{lin2014microsoft} to general domains such as LAION-5B \cite{schuhmann2022laion5b}. This evolution is driven by the emergence of large vision-language models (VLMs) like CLIP \cite{CLIP} and ALIGN \cite{jia2021scaling}. Leveraging these models and datasets, various approaches have been introduced, including auto-regressive models like DALL-E \cite{DALLE2}, CogView \cite{ding2021cogview}, and Parti \cite{yu2022scaling}; mask-based transformers such as MUSE \cite{chang2023muse} and MaskGIT \cite{chang2022maskgit}; GAN-based models such as StyleGAN-T \cite{sauer2023stylegan}, GigaGAN \cite{kang2023scaling}; and diffusion models like GLIDE \cite{Nichol2021GLIDETP}, Imagen \cite{Saharia2022PhotorealisticTD}, Stable Diffusion (SD) \cite{SD}, DALL-E2 \cite{DALLE2}, DALL-E3 \cite{DALLE3}, and eDiff-I \cite{ediff}. Among these, diffusion models are popular due to their ability to generate high-quality images. 
However, they typically require many-step sampling to generate high-quality images, limiting their real-time and on-device applications.

\subsection{Accelerating Text-to-Image Diffusion Models} 
Efforts to accelerate diffusion model sampling include faster samplers and distillation techniques. Early methods reduce sampling steps to as few as 4-8 steps by incorporating Latent Consistency Models to distill latent diffusion models \cite{LCM, pixart}. Recent studies have achieved one-step text-to-image generation by training a student model distilled from a pretrained multi-step diffusion model, employing various techniques such as Rectified Flows \cite{InstaFlow}, reconstruction and distribution matching losses \cite{DMD}, and adversarial objectives \cite{SDTurbo,lin2024sdxllightning,ufo, mobilediffusion}. However, the output images often exhibit blurriness and artifacts, and one-step methods still underperform compared to multi-step models while requiring large-scale text-image pairs for training.


Differentiating itself from the rest, SwiftBrush \cite{SwiftBrush} proposed a one-step distillation technique that required only training on prompt inputs. The method gradually transfers knowledge from the teacher to the one-step student through an intermediate LoRA multi-step teacher. SwiftBrush's image-free training procedure offers a simple way to scale training data and extend the student model capability via auxiliary losses, which are not constrained by limited-size imagery training data. Therefore, while SwiftBrush also falls short in quality compared to the teacher model, we find its high potential for further development to produce a one-step student that even beat the multi-step teacher at its own game.




\section{Background}

\myheading{Diffusion Models} are generative models that transform a noise distribution into a target data distribution by simulating the diffusion process. This transformation involves gradually adding noise $\epsilon \sim \mathcal{N}(0, I)$ to clean image $\mathbf{x}_0$ over a series of $T$ steps (forward process) and then learning to reverse this process (reverse process). The forward process can be formulated as:
\begin{equation}
    \mathbf{x}_t = \alpha_t \mathbf{x}_0 + \sigma_t \epsilon \quad \forall t \in \overline{0,T}
\end{equation}
where $\mathbf{x}_t$ is the data at time step $t$ and $\{(\alpha_t, \sigma_t) \}_{t=1}^T$ is the noise schedule such that $(\alpha_T,\sigma_T)=(0,1)$ and $(\alpha_0, \sigma_0)=(1,0)$. 
On the other hand, the reverse process aims to reconstruct the original data from noise. Training involves minimizing the difference between predicted output from model $\epsilon_\phi$ parameterized by $\phi$ and the actual added noise:
\begin{equation}\label{eq:loss_lora}
    \min_{\phi} \mathbb{E}_{t \sim \mathcal{U}(0, T),   \epsilon \sim \mathcal{N}(0, I)} \Vert \epsilon_{\phi}(\mathbf{x}_t, t) - \epsilon \Vert_2^2
\end{equation}

\myheading{Text-to-Image Diffusion Models} generate images from textual descriptions by integrating text embeddings within the inference process, aiming to align textual descriptions with visual outputs. A key challenge is the lack of mechanisms to ensure textual relevance and image fidelity.
To deal with this, large-scale text-to-image diffusion models usually utilize classifier-free guidance which enhances text-image alignment without a separate classifier. This is achieved by interpolating the model's output with and without the conditioning text, controlled by a guidance scale $\gamma$. The model's final output with guidance is given by:
\begin{equation}
    \hat{\epsilon}_\phi(\mathbf{x}_t, t, \mathbf{y}) = \epsilon_\phi(\mathbf{x}_t, t, \mathbf{y}) + \gamma \cdot (\epsilon_\phi(\mathbf{x}_t, t, \mathbf{y}) - \epsilon_\phi(\mathbf{x}_t, t)),
\end{equation}
where $\epsilon_\phi(\mathbf{x}_t, t, \mathbf{y})$ is the predicted output conditioned on text embedding $\mathbf{y}$ and $\epsilon_\phi(\mathbf{x}_t, t)$ is the unconditional prediction, i.e., using null text. This formula showcases how classifier-free guidance directly influences the generation process, leading to more precise and relevant image outputs.

\myheading{SwiftBrush} \cite{SwiftBrush} is a one-step text-to-image generative model, employing an image-free distillation technique inspired by text-to-3D synthesis methods. At the heart of this approach is re-purposing Variational Score Distillation (VSD)\cite{wang2023prolificdreamer}, a novel loss that tackles the challenges of over-smoothing and diversity reduction often seen in early text-to-3D work \cite{poole2022dreamfusion}. Specifically, SwiftBrush uses two teachers, one frozen teacher $\epsilon_\phi$ and one LoRA \cite{hu2021lora} teacher  $\epsilon_\psi$, to guide the one-step student $f_\theta$. Here, the LoRA teacher is to bridge the gap between the frozen teacher and the student. On one hand, the loss for training the student model is formalized as:
\begin{equation}
    \nabla_{\theta} \mathcal{L}_{VSD} = \mathbb{E}_{t, \mathbf{y}, \mathbf{z}} \left[ w(t) (\hat{\epsilon}_{\phi}(\mathbf{x}_t, t, \mathbf{y}) - \hat{\epsilon}_{\psi}(\mathbf{x}_t, t, \mathbf{y})) \frac{\partial f_{\theta}(\mathbf{z}, \mathbf{y})}{\partial \theta} \right],
\end{equation}
where $\mathbf{z} \sim \mathcal{N}(0, I)$ is the input noise to the student network, $\mathbf{x}_t = \alpha_t \hat{\mathbf{x}}_0 + \sigma_t \epsilon$ is the noise-added version at a time step $t$ of the student's output $\hat{\mathbf{x}}_0 = f_{\theta}(\mathbf{z},y)$, and $w(t)$ is the weighting of the loss. On the other hand, the LoRA teacher is trained using diffusion loss $\|\mathbb{E}_{t,\epsilon,y}[\epsilon_{\psi}(\mathbf{x}_t, t, \mathbf{y}) - \epsilon]\|^2_2$. SwiftBrush alternates between the training of the student model and the LoRA teacher until convergence.


\section{Proposed Methods}
\begin{figure}[t]
    \centering
    \includegraphics[width=.96\textwidth]{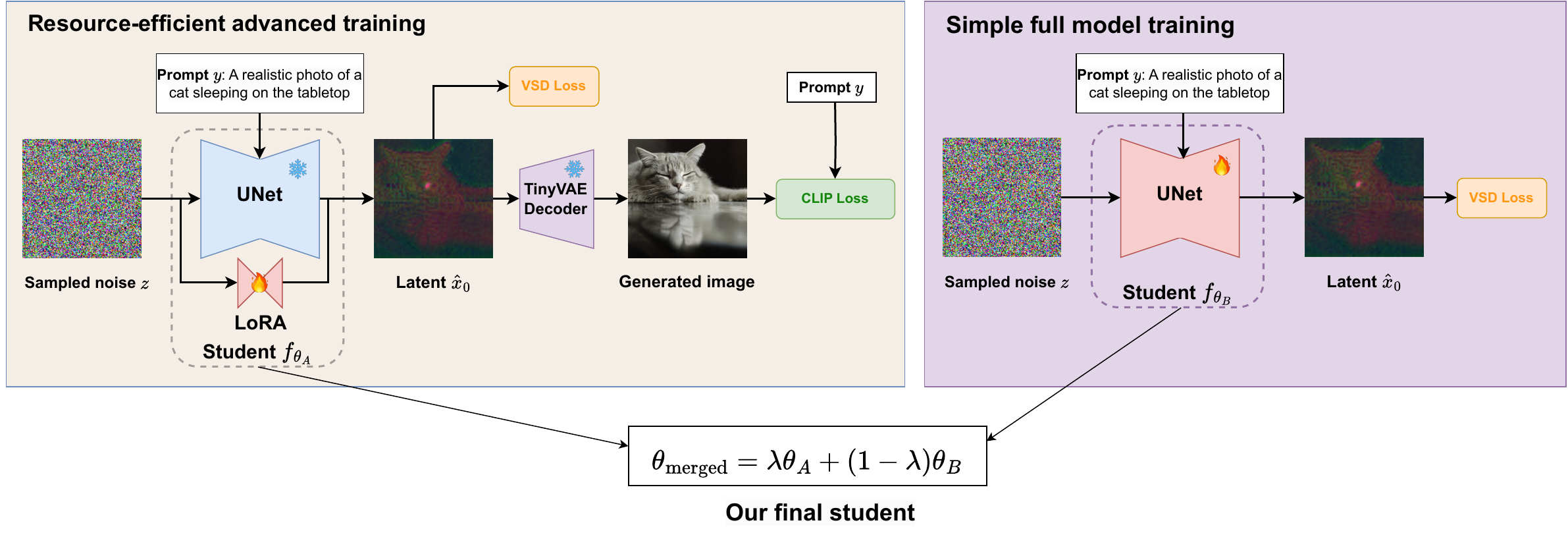}
    \caption{\textbf{SwiftBrush v2 overview:} two versions of the student model: a fully finetuned model trained with the Variational Score Distillation (VSD) loss, and a LoRA finetuned model trained with both VSD and CLIP loss. The final model is obtained by merging the two student models, leveraging the strengths of both training schemes.}
    \label{fig:systemfig}
\end{figure}

In this section, we begin by conducting an in-depth analysis of the quality-diversity trade-off in representative diffusion-based text-to-image models (\cref{sec:analysis}). Subsequently, we discuss our strategy for incorporating the strengths of SwiftBrush and SD Turbo (\cref{sec:integration}). Lastly, we explore various approaches to enhance the distillation process and post-training procedures (\cref{sec:intraining} to \ref{sec:posttraining}). An overview of our methodologies is presented in \cref{fig:systemfig}.
\subsection{Quality-Diversity Trade-off in Existing Models}
\label{sec:analysis}

We first analyze the properties of the teacher model, SDv2.1, and existing one-step diffusion-based text-to-image models. For the teacher model, we assess its performance across different guidance scales. 
Here, we select SwiftBrush and SD Turbo due to their quality and distinct training procedures.
 SwiftBrush relies solely on score distillation from the teacher in its image-free training, while SD Turbo trains on real images with adversarial and distillation loss.
We conduct our analysis on the COCO 2014 benchmark and report relevant metrics in \cref{tab:analysis}.

\begin{table}[t]
\centering
\setlength{\tabcolsep}{5pt}
\caption{Comparison between the multi-step teacher (SDv2.1), SD Turbo, and SwiftBrush on the zero-shot MS COCO-2014 30K benchmark. The best scores are in \textbf{bold}, while the better scores among one-step models are \underline{underlined}.}
\label{tab:analysis}
\begin{tabular}{lccccc}
\toprule
\textbf{Model Name} & \textbf{NFE} & \textbf{FID$\downarrow$}  & \textbf{CLIP$\uparrow$} & \textbf{Precision$\uparrow$}    & \textbf{Recall$\uparrow$}  \\ 
\midrule
SD 2.1 (\textit{cfg} = 2)    & 25           & \textbf{9.64} & 0.31             & 0.57                  & 0.53             \\
SD 2.1 (\textit{cfg} = 4.5)  & 25           & 12.26         & \textbf{0.33}          & 0.61                  & 0.41             \\
SD 2.1 (\textit{cfg} = 7.5)  & 25           & 15.93         & \textbf{0.33}          & 0.59                  & 0.36             \\ 
\midrule
SD Turbo            & 1            & 16.10         & \underline{\textbf{0.33}} &  \textbf{\underline{0.65}} & 0.35             \\
SwiftBrush          & 1            & \underline{15.46} & 0.30                    & 0.47                  & \underline{0.46}    \\
\bottomrule

\end{tabular}
\end{table}

When assessing the multi-step teacher's performance, the classifier-free guidance scale (\textit{cfg}) plays a crucial role. A low \textit{cfg} (e.g., \textit{cfg} = 2) yields a low FID score of 9.64, driven by high output diversity (recall = 0.53). However, this setting results in weak alignment between images and prompts (CLIP score = 0.30) and lower image quality (low precision). Conversely, a large \textit{cfg} (e.g., \textit{cfg} = 7.5) markedly improves text-image alignment (CLIP score = 0.33) but restricts diversity (recall = 0.36), resulting in a poor FID score of 15.93. Moderate \textit{cfg} values (e.g., \textit{cfg} = 4.5) strike a better balance, offering the highest precision score.

When evaluating the one-step students, we notice distinct behaviors. SD Turbo, benefiting from adversarial training on real images, yields highly naturalistic outputs with an exceptionally high precision score, surpassing even those of the multi-step teacher. However, this results in poor diversity, reflected in a low recall of 0.35. Conversely, SwiftBrush adopts an image-free training approach, allowing flexible combinations of random-noise latents and input prompts. Such a relaxed supervision enables the student model to generate more diverse outputs but at the expense of quality (\cref{tab:analysis}). We further verify this finding by a qualitative evaluation, illustrated in \cref{fig:teaser}. When given identical input prompts, SD Turbo generates realistic yet similar outputs. In contrast, SwiftBrush produces a wider range of outcomes, albeit with greatly distorted artifacts. Regardless, both one-step models exhibit FID scores around 15-16, significantly higher than the teacher's best score. Observing a quality-diversity trade-off in existing one-step diffusion model such as SD Turbo and SwiftBrush, we aim to combine these two to leverage the strengths of both.

\subsection{SwiftBrush and SD Turbo Integration}
\label{sec:integration}
In this section, we explore strategies for effectively merging SwiftBrush and SD Turbo to enhance the quality-diversity trade-off. A direct approach is unifying their training procedure, i.e., combining adversarial training from SD Turbo and Variational Score Distillation from SwiftBrush. However, this simplistic approach proves challenging due to computational demands and potential failure. While SwiftBrush's image-free procedure is easier to implement, reproducing SD Turbo's training process is complex and resource-intensive. The presence of the discriminator complicates the training, necessitating significant VRAM and dataset requirements. Additionally, SD Turbo's intense supervision may constrain SwiftBrush's loose guidance, limiting output diversity.

Based on our discussions, we opt not to utilize SD Turbo's adversarial training. Instead, we leverage its pretrained weights to initialize the student network within SwiftBrush's training framework. This straightforward approach proves highly effective. As can be seen in the comparison between the second and the third row in \cref{tab:datasetcompare}, the resulting model has improved FID and recall. By employing SD Turbo's pretrained weights, we provide a solid foundation for the training model to maintain high-quality outputs, while SwiftBrush's image-free training process gradually enhances generation diversity.


\subsection{In-training Improvements}
\label{sec:intraining}
Besides data efficiency and diversity promotion, SwiftBrush's image-free training still has room for improvement. First, it allows an easy means to scale up training data by collecting more prompt inputs. This task is simple, given the abundance of textual datasets and the availability of large language models, unlike the costly and labor-intensive task of collecting image-text pair data commonly required. Second, by not forcing the output of the student model to be the same as that of the teacher, SwiftBrush allows the student to go even beyond the quality and capability of the teacher. We can advocate it to happen by adding extra auxiliary loss functions in SwiftBrush training. In this section, we will discuss the implementation of those ideas for improvement.

\myheading{Implications of Dataset Size.} 
SwiftBrush's image-free approach allows for scalable training datasets without limitations. To explore the dataset's impact on SwiftBrush performance, we conducted supplementary experiments by augmenting the dataset with an additional 2M prompts from the LAION dataset \cite{schuhmann2022laion5b} to the original 1.5M deduplicated prompts from the JourneyDB dataset \cite{pan2023journeydb}. Analysis (\cref{tab:datasetcompare}) reveals improved performance with the expanded dataset. Specifically, this leads to a significant improvement in terms of FID and precision, suggesting a positive correlation between dataset size and the quality of the generated outputs. However, a slight degradation in recall was observed, indicating a potential trade-off between image diversity and overall quality. Furthermore, despite an increase in CLIP score compared to the previous version, there remains room for improvements in terms of text alignment. 


\input{tables/datasetcompare}

\myheading{Tackling the text-alignment problem.}
\label{sec:text_align}
To refine the coherence between textual prompts and visual outputs, we integrate an additional CLIP loss within the distillation process.
However, naively employing such loss between the student model's predictions and the original textual prompts poses challenges
, as over-optimizing for the CLIP score potentially degrade image quality. We observed issues such as blurriness, increased color saturation, and the emergence of textual artifacts within the generated images.

To address this, we propose clamping the CLIP value during training with ReLU activation. This aims to balance text alignment with preserving image quality, ensuring the model maintains visual integrity. Additionally, we introduce dynamic scheduling to control the influence of CLIP loss, gradually reducing its weight to zero by the end of distillation. This balanced approach integrates visual-textual alignment and image fidelity effectively. Our clamped CLIP loss is formulated as:
\begin{equation}
\mathcal{L_{\text{CLIP}}}=\max \left(0, \tau - \langle \mathcal{E}_{\text{image}}\left(\mathcal{D}\left(f_{\theta}(\mathbf{z}, \mathbf{y})\right)\right), \mathcal{E}_{\text{text}}(\mathbf{y})\rangle\right),\label{eq:clip_loss}
\end{equation}
where $\mathcal{E}_{\text{image}}$ and $\mathcal{E}_{\text{text}}$ represent the CLIP image and text encoders, respectively. $\mathcal{D}$ is the VAE decoder used to map the latent back to the image. The term $\tau$ introduces a threshold on the desired cosine similarity $\langle\cdot,\cdot\rangle$ between the image and text embeddings, preventing the model from overemphasizing textual alignment at the expense of image quality.

\subsection{Resource-efficient training schemes}\label{sec:efficient}
While our CLIP loss is highly beneficial, it comes with memory and computation costs. Particularly, the CLIP image encoder can only work on image space, requiring decoding the predicted latent to image via the image decoder $\mathcal{D}$ as can be seen in \cref{eq:clip_loss}. We find incorporating CLIP loss into SwiftBrush's full-model distillation significantly slows down training speed, particularly on GPUs with moderate VRAM. This urges us to design a resource-efficient training scheme to fully exploit the proposed CLIP loss in constrained setting.

It is possible to significantly reduce memory requirements during fine-tuning with the LoRA framework \cite{hu2021lora}, where only a set of small-rank parameters are trained. Also, to compute the CLIP loss, the predicted latent goes through a large VAE decoder, increasing training length and memory consumption. To address this, we integrate TinyVAE \cite{tinyaesd}, a compact variant of Stable Diffusion's VAE. TinyVAE sacrifices some fine detail in images but preserves overall structure and object identity comparable to the original VAE. This approach maintains training efficiency close to those of the original fully fine-tuned model, as shown in \cref{tab:clipcompare,sec:ablate_efficient}.

\subsection{Post-training improvements}
\label{sec:posttraining}
\begin{figure}[t]
    \centering
    \includegraphics[width=.9\linewidth]{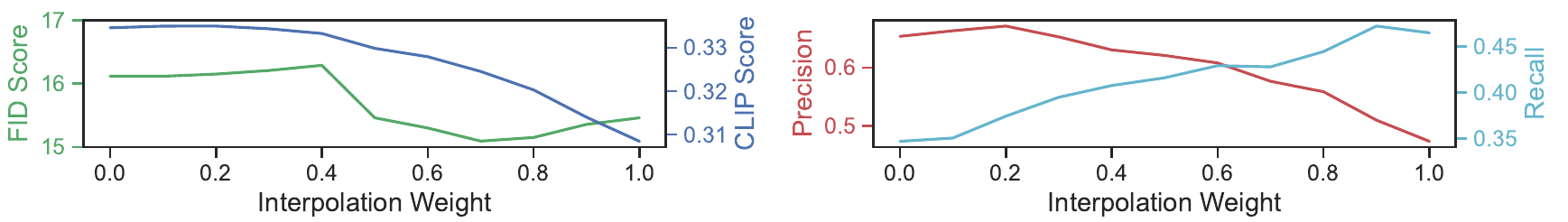}
    \caption{\textbf{The effect of weight interpolation} upon FID, CLIP score, precision, and recall calculated on the zero-shot MS COCO-2014 benchmark. 0.0 indicates SD Turbo, and 1.0 indicates the original SwiftBrush.}
    \label{fig:turbo_sb_interp}
\end{figure}

Recent literature \cite{deepmodelfusionsurvey} has shown a growing interest in model fusion techniques, aiming to integrate models performing distinct subtasks into a unified multitask model \cite{jin2022dataless, tiesmerging} or combining fine-tuned iterations to create an enhanced version \cite{modelsoups, izmailov2018averaging, choshen2022fusing}. Our research focuses on the latter, particularly in the context of one-step text-to-image diffusion models. These models, although designed for the same task, differ in their training objectives, providing each with unique advantages. By merging these models, we aim to create a new model that captures the strength of each model without increasing model size or inference costs. Given two one-step diffusion models with weights $\theta_{A}$ and $\theta_{B}$ and an interpolation weight $\lambda$, we merge them using a simple linear interpolation of the weights:
\vspace{-0.5mm}
\begin{equation}
    \theta_{\text{merged}} = \lambda \theta_{A} + (1-\lambda) \theta_{B}.
\end{equation}
\vspace{-0.5mm}
We empirically demonstrate the benefit of such interpolation scheme with SD Turbo, known for its precision and strong text alignment, and the original SwiftBrush, which excels in diversity. In our empirical analysis (refer to \cref{fig:turbo_sb_interp}), we observe that by interpolating from one model to the other, all evaluated metrics (except for the CLIP score) show improvement at some optimal point. This indicates that the fused model could potentially outperform the original models. These findings underscore the potential of model fusion techniques in enhancing model efficacy, as evidenced by the metric analysis.

As discussed in \cref{sec:efficient}, our proposal suggests two training schemes. We can either train the student model with LoRA and TinyVAE utilizing VSD and CLIP losses or fully finetune the student model employing only VSD loss. These two training schemes lead to two resulting one-step models with different behaviors, making them ideal ingredients for merging. By merging these models, we obtain the final model output of our proposed SwiftBrush v2 framework.

\section{Experiments}
\subsection{Experimental Setup}

\myheading{Evaluation metrics.} Our text-to-image model is evaluated using the \textbf{``zero-shot''} setting, i.e., trained on some datasets and tested on another dataset, undergoing comprehensive evaluation across three key aspects: image quality, diversity, and textual fidelity. We use the Fréchet Inception Distance (FID)\cite{fid} on resized $256 \times 256$ images as our primary metric for evaluating image quality, consistent with prior text-to-image research\cite{kang2023gigagan}. In addition, we employ precision \cite{improvedrecall} as a complementary metric to FID. For evaluating diversity, we rely on the recall metric \cite{improvedrecall}. Textual alignment is measured using the CLIP \cite{CLIP} score and the Human Preference Score v2 \cite{wu2023human} (HPSv2).

\myheading{Datasets.} 
We utilize two training datasets: (1) 1.3M prompts from JourneyDB \cite{pan2023journeydb}, and (2) an expanded dataset incorporating 2M prompts from LAION \cite{schuhmann2022laion5b}, totaling 3.3M prompts. Additionally, a human-feedback dataset, comprising 200K pairs from LAION-Aesthetic-6.25+ \cite{schuhmann2022laion5b}, can be optionally used for further image regularization \cite{DMD}, which is around 5\% of the total training data.
We use the MS COCO-2014 validation set as the standard zero-shot text-to-image benchmark, consistent with established practices in the field \cite{InstaFlow,BOOT,SDTurbo,LCM,sauer2023stylegan,kang2023gigagan}. Samples are generated from the first 30k prompts, with the entire dataset serving as the reference for obtaining metrics. For HPSv2, we adopt the evaluation protocol from \cite{wu2023human}.

\myheading{Training details.} 
Our method is built on top of SwiftBrush \cite{SwiftBrush} with our proposed modifications. We conduct all our training on four NVIDIA A100 40GB GPUs, with training durations of one or three days depending on the dataset (JourneyDB alone or combined with LAION prompts). The batch size is 16 per GPU, and we use learning rates of $1e^{-6}$ and $1e^{-3}$ for the student and LoRA teacher, respectively, with the AdamW optimizer \cite{adamw}. Our approach utilizes Stable Diffusion 2.1 \cite{SD} with $\textit{cfg}=4.5$ as the frozen teacher and LoRA teacher initialized
with rank $r=64$ and scaling $\gamma=128$. As for the LoRA student, we set $r=256$ and $\gamma=512$ to enhance its learning capacity. In addition, we introduce the clamped CLIP loss with a margin of $\tau=0.35$, starting with a weight of 0.1 and gradually reducing to zero. We use ViT-B/32 \cite{openclip} as the backbone for CLIP image and text feature extraction. Finally, we merge two final models with $\lambda=0.5$, further details are available in the Appendix.

\input{tables/work_pr_compare}
\input{tables/hps}

\subsection{Comparison with Prior Approaches}
\myheading{Quantitative results.}
\cref{tab:coco_compare} presents a comprehensive quantitative comparison between our approach and prior text-to-image models. This encompasses GAN-based models (group 1), multi-step diffusion models (group 2), and a variety of distillation techniques (group 3), both with and without image supervision. Our approach outperforms all competitors, notably achieving superior results even without direct image regularization. Remarkably, our distilled student models exceed their teacher model, SDv2.1, in FID scores by a significant margin while maintaining equivalent model size and inference times comparable to SD Turbo or SwiftBrush. Our model effectively addresses previous text alignment issues observed in SwiftBrush, being close to the CLIP scores of SD Turbo and multi-step models. Precision metrics show high-realism akin to the reference dataset, enhanced solely with a text-driven training dataset. Notably, ours exhibits significant recall improvements due to its image-free nature. Image-based regularization further improves student quality with a small reduction in recall. 

In terms of HPSv2 (\cref{tab:hps_compare}), our approach achieves competitive scores compared to the multi-step teacher model SDv2.1 and other distillation methods. Particularly, our model with image regularization achieves the highest HPSv2 scores for photos and remains close to the top performers in other categories. 

\begin{figure}[t]
    \centering
    \includegraphics[width=.85\textwidth]{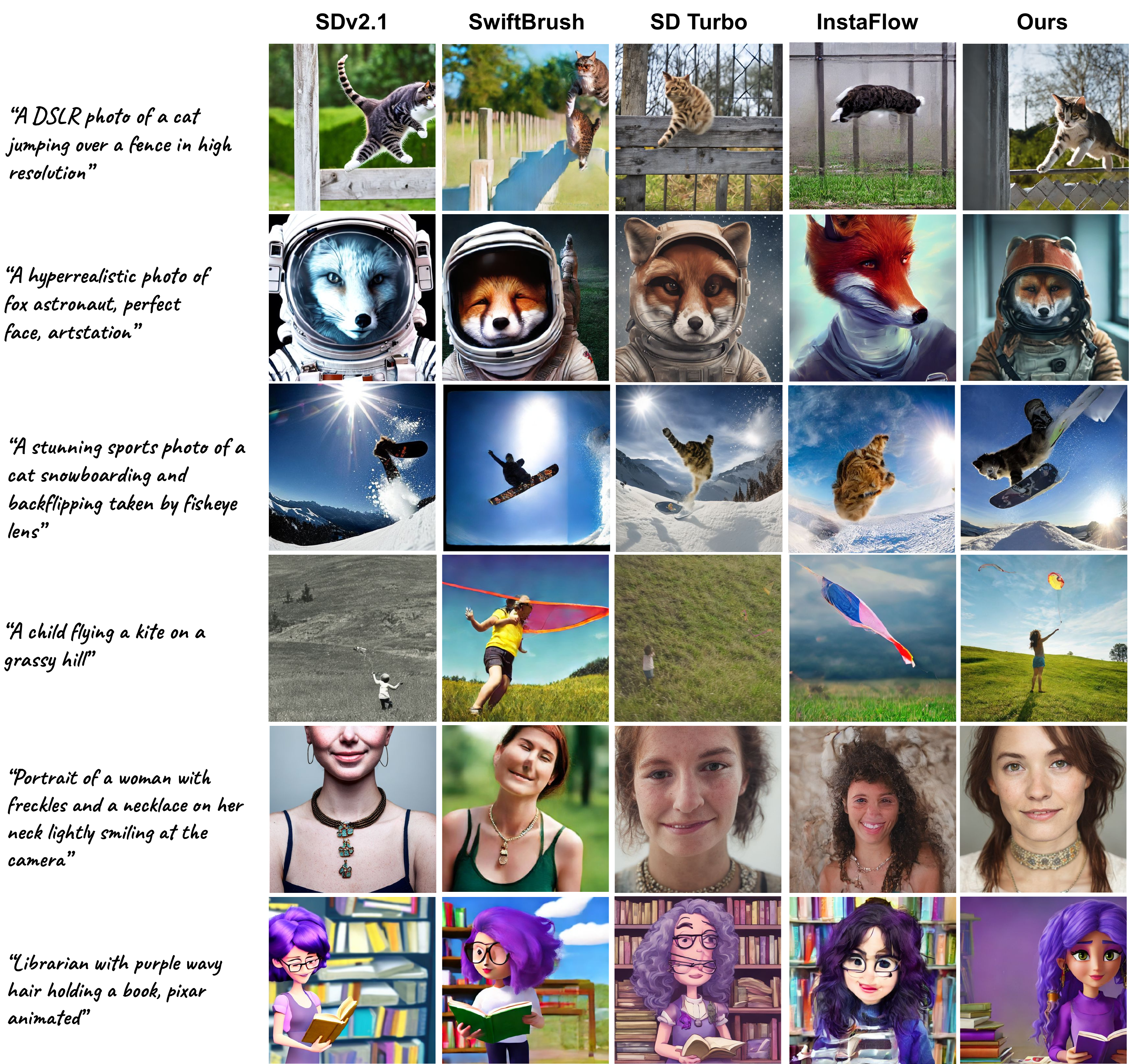}
    \caption{Exemplified images generated by SD Turbo, SwiftBrush, SDv2.1 with 50 sampling steps, InstaFlow-0.9B and Ours. Images in the same row are sampled from the same text prompt, while images in the same column are from the same model.}
    \label{fig:quality}
\end{figure}

\myheading{Qualitative results.}
We provide a qualitative comparison in \cref{fig:quality}. Our model produces higher quality compared to its teachers and one-step counterparts. In the first row, despite correctly illustrating the jumping action, other models yield deformed shapes of cats. This issue also reappears in the third row. In contrast, our model realistically represents the action and the cat's figure. SDv2.1 encounters out-of-frame issues in the fourth row, while others generate facial deformities in the last two rows. Meanwhile, our model produces artifact-free faces that are aligned with the text description regarding hairstyle and expression. 

\begin{wrapfigure}{r}{0.54\textwidth}
    \centering
    \includegraphics[width=0.53\textwidth]{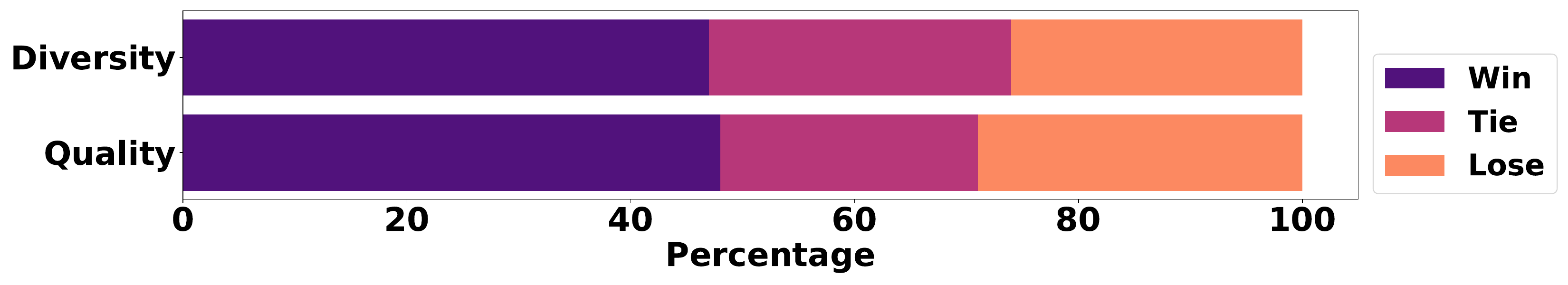}
    \caption{\textbf{User survey.} We asked participants to compare the quality and diversity of images generated by our method and its teacher model across 20 random text prompts.}
    \label{fig:survey}
\end{wrapfigure}

\begin{figure}[t!]
    \centering
    \includegraphics[width=.85\textwidth]{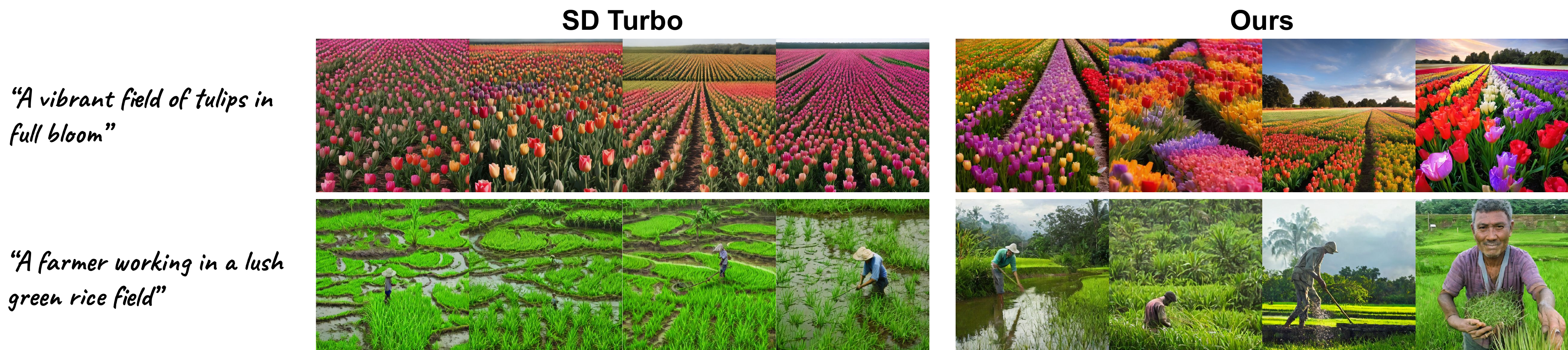}
    \caption{Exemplified images generated by SD Turbo and Ours to demonstrate our method's high diversity.}
    \label{fig:diversity}
\end{figure}

Among the counterparts, SD Turbo shows the best image quality but also the worst diversity. We reaffirm its weakness when comparing it to our method in \cref{fig:diversity}. SD Turbo tends to generate similar images, as evidenced by the similar details and colors for different samples in each prompt. In contrast, our model can generate diverse views and colors in the first prompt as well as different environments in the second prompt.

We also surveyed 250 participants to compare our distilled model with its multi-step teacher, SDv2.1. Participants evaluated images generated by each model for 20 random prompts, and the results (\cref{fig:survey}) show that our method consistently matched or exceeded the teacher in quality and diversity.



\input{tables/config}
\input{tables/clipcompare}

\subsection{Ablation Studies}

\myheading{Effect of each proposed component} is summarized in \cref{tab:ablation}. We compare two student training schemes: (1) full model training and (2) efficient model training. Initializing the model from SD Turbo \cite{SDTurbo} significantly improves the FID in both schemes. Adding prompts from LAION \cite{schuhmann2022laion5b} leads to notable FID improvements in both cases, reaching 11.27 in (1) and 11.02 in (2). Also, efficient training allows for the use of the Clamped CLIP loss, resulting in a considerable FID improvement. Notably, combining models from both schemes further reduces the FID to 8.77, surpassing their SDv2.1 teacher. Finally, with extra image regularization, we achieve a further boost, reaching an FID of 8.14.




\myheading{Detailed study in clamped CLIP loss} is shown in 
\label{sec:ablate_efficient}
\cref{tab:clipcompare}. Naively applying CLIP loss \cite{CLIP} worsens performance, increasing the FID by 5 points. In contrast, our proposed clamped CLIP loss (\cref{eq:clip_loss}) is highly effective, reducing the FID to 13.19. Moreover, employing a weight scheduler for the clamped CLIP loss further enhances performance, resulting in an FID of 11.70.

\section{Discussion and Conclusion}
\myheading{Limitations:}
Despite the promising results, our distilled model still inherits some limitations from the teacher model, such as compositional problems. To address these limitations, future work could explore the integration of auxiliary losses that focus on cross-attention mechanisms during the distillation process.


\myheading{Societal Impact:}
Our advancements improve high-quality image synthesis speed and accessibility. However, misuse of our advancements could spread misinformation and manipulate public perception. Thus, responsible use and safeguards are crucial to ensure that the benefits outweigh the risks.


\myheading{Conclusion:}
This paper proposes a novel method to enhance SwiftBrush, a one-step text-to-image diffusion models.
We address the quality-diversity trade-off by initializing the SwiftBrush student model with SD Turbo's pretrained weights and incorporating efficient in-training techniques as well as margin CLIP loss and large-scale dataset training. Also, by weight merging and optional image regularization, we achieve an outstanding FID score of 8.14, surpassing existing approaches in both GAN-based and one-step diffusion-based text-to-image generation while maintaining near real-time inference speed.



%
%
\bibliographystyle{splncs04}
\bibliography{main}

\maketitlesupplementary

\begin{figure}[b]
    \centering
    \begin{subfigure}[b]{\textwidth}
        \centering
        \includegraphics[width=\textwidth]{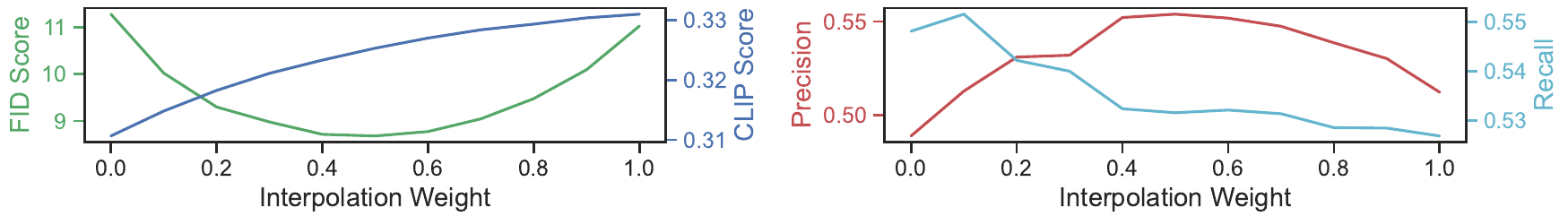}
        \subcaption{Model \textbf{Ours}.}
    \end{subfigure}
    \begin{subfigure}[b]{\textwidth}
        \centering  
        \includegraphics[width=\textwidth]{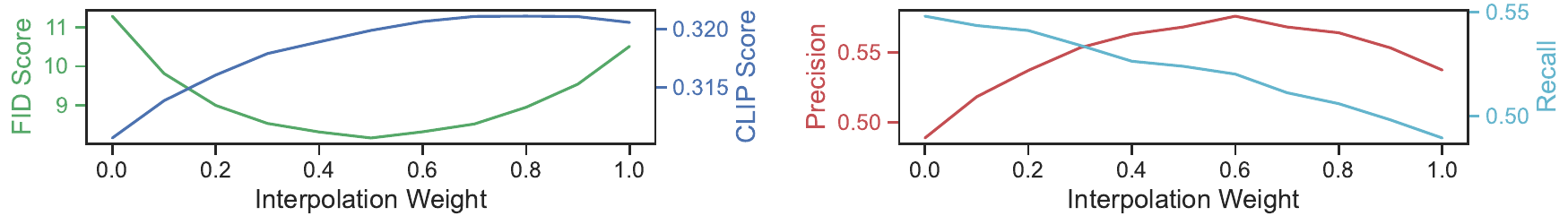}
        \subcaption{Model \textbf{Ours*.}}
    \end{subfigure}
    \caption{\textbf{The effect of weight interpolation} upon FID, CLIP score, precision, and recall calculated on the zero-shot MS COCO-2014 benchmark when combining models trained with our proposed mechanisms (Model A and B) to form the final model.}
    \label{fig:compare_interp}
\end{figure}

\section{Additional details}
\subsection{Weight Interpolation}
In this section, we first provide quantitative analyses on the model merging process conducted on the fully finetuned (Model A) and the resource-efficient trained model (Model B) to form our final model. We run a comprehensive evaluation by reporting essential metrics, including FID, CLIP score, Precision, and Recall, upon the zero-shot MS-COCO 2014 across different interpolation weights, following the same protocol as in the main paper. We provide the plots in both scenarios when the regularization term is applied  (\cref{fig:compare_interp}.b) or not (\cref{fig:compare_interp}.a). In either case, we observe that the CLIP and the precision scores change monotonically from one model to another, while both the FID and recall scores get enhanced when fusing the two models. This analysis provides a data-driven justification for the selected interpolation weight used in the final model, ensuring it achieves the best combination of visual quality, semantic coherence, and diversity. Specifically, for both cases, we pick the weight to optimize for the FID metrics, while not trading off too much with other metrics, hence the interpolation weight $\lambda=0.5$ serve well with our purposes.

\begin{figure}[t]
    \centering
    \begin{subfigure}[b]{\textwidth}
        \centering
        \includegraphics[width=\textwidth]{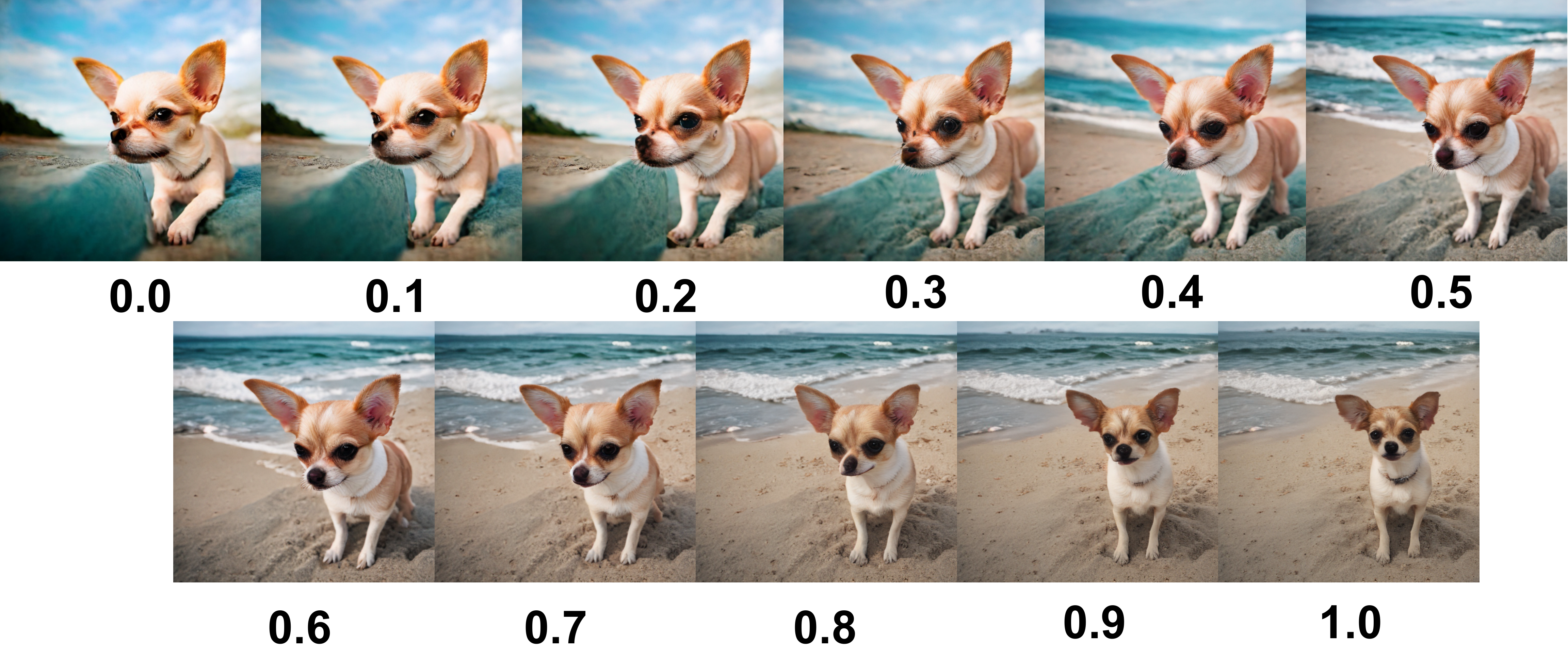}
        \subcaption{Interpolation of original SwiftBrush and SD Turbo. Prompt: ``A chihuahua at the beach''}
    \end{subfigure}
    \begin{subfigure}[b]{\textwidth}
        \centering
        \includegraphics[width=\textwidth]{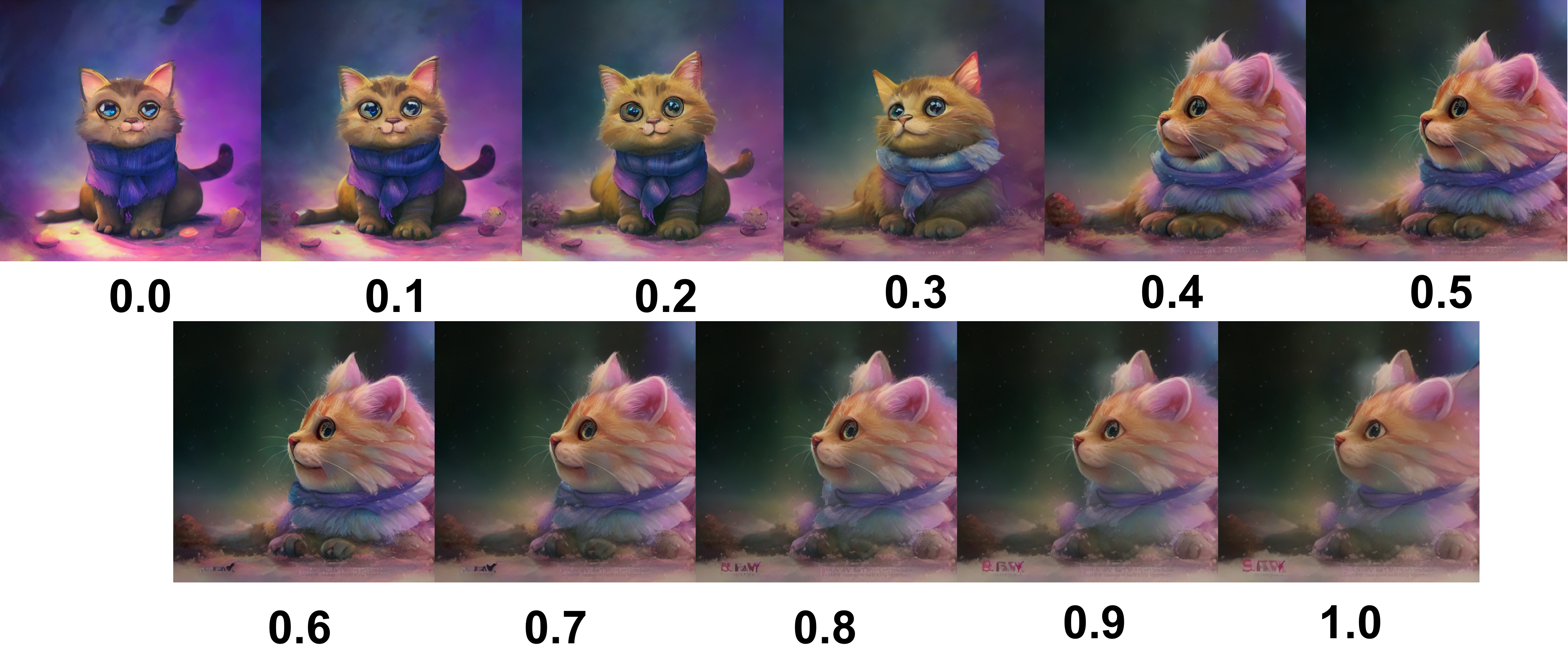}
        \subcaption{Model Ours* (combined between model A - fully fine-tuned and model B - efficient fine-tuned with LoRA and CLIP loss). Prompt: ``cat, airbrush style, soft lighting, detailed face, snowy, concept art, digital painting, epic.''.}
    \end{subfigure}
    \caption{\textbf{The effect of weight interpolation} upon the generated samples.}
    \label{fig:compare_interp_sample}
\end{figure}

Furthermore, we delve into the visual analysis of model interpolation (\cref{fig:compare_interp_sample}), exploring the effects on the generated output as the interpolation weight is varied between two trained one-step text-to-image models. We provide qualitative figures for both the interpolation between SwiftBrush and SD Turbo (analyzed in the main paper) and the interpolation between Model A and B mentioned above. By gradually adjusting the interpolation weight, we can observe how the visual characteristics of the generated samples evolve, providing insights into the learned representations of each model and their contributions to the final output. This interpolation study allows us to understand the interplay between the two models and how their combined representation affects the generated results.


\begin{figure}[t]
    \centering
    \includegraphics[width=\linewidth]{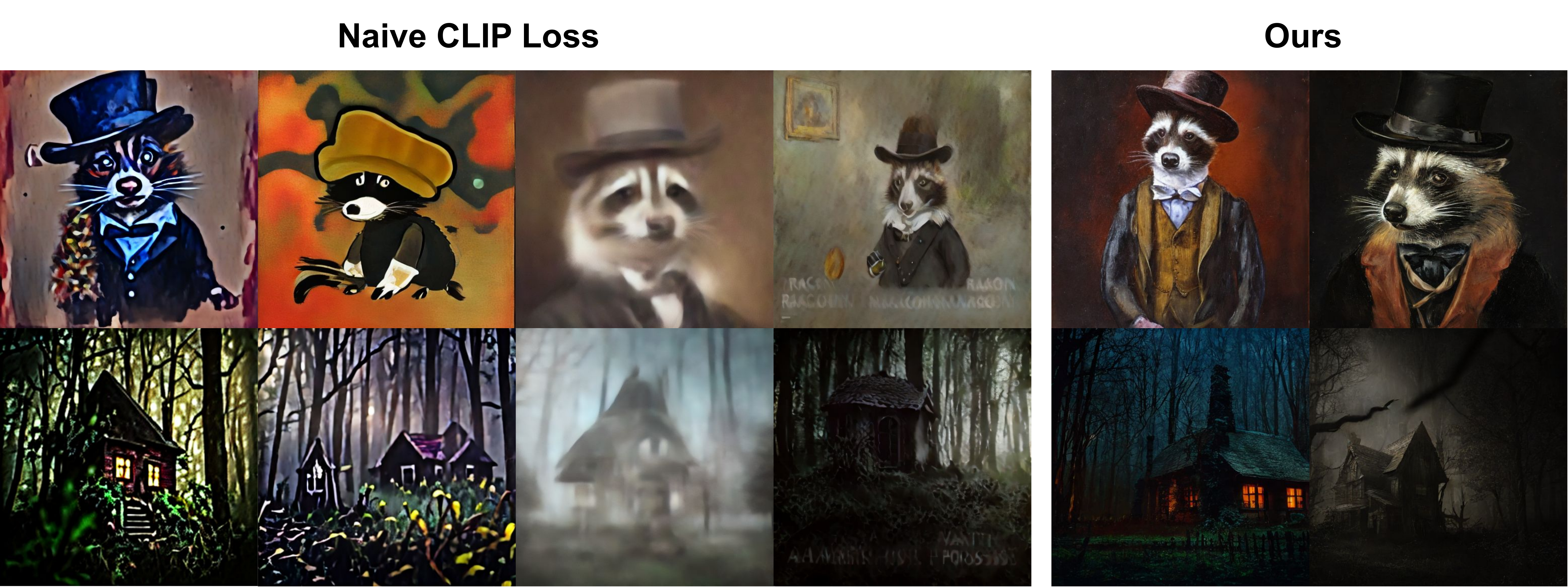}
    \caption{\textbf{Effect of the CLIP loss design.} We provide a qualitative comparison between naively adding CLIP loss and our approach. The output of the first approach is over-smooth, over-saturated, and has textual artifacts on the image. On the other hand, our approach shows both good diversity and quality. The prompt for the first row is: ``A raccoon wearing formal clothes, wearing a tophat. Oil painting in the style of Rembrandt'', while the second's is: ``a beautiful dark forest, a witches house in the trees, creepy, dark wooded''.}
    \label{fig:badclip}
\end{figure}

\subsection{CLIP loss}
\cref{fig:badclip} presents a qualitative comparison between the naive approach of integrating the CLIP loss and our proposed method. The output of the distilled model using the naive approach suffers from poor quality, with issues such as over-saturation, over-smoothing, and the appearance of textual artifacts on the image that reflect the conditioned prompt. These artifacts can be visually distracting and detract from the overall aesthetic of the generated images. In contrast, when the loss is applied properly using our method, the generated images exhibit significant improvements in both quality and diversity. The images are more visually appealing, with better color balance, sharper details, and a greater range of content that accurately captures the essence of the input prompt. By carefully integrating the CLIP loss into the distillation process, our approach effectively leverages the semantic understanding of the CLIP model while preserving the generative capabilities of the underlying model. This results in high-quality, diverse images that closely align with the desired output specified by the prompt.

\subsection{Dependency on the existing one-step diffusion   }
\begin{figure}[h]
    \centering
    \begin{minipage}{0.5\linewidth}
        \centering
        \footnotesize
        \captionof{table}{Comparison between SwiftBrush (SB) and our method when using SB as initialization weight.}
        \label{tab:rebuttal}
        \begin{tabular}{lccc}
            \toprule
            \textbf{Approach} & \textbf{FID} & \textbf{Pre} & \textbf{Rec} \\ 
            \midrule
            SB & 15.46 & 0.47 & 0.46 \\
            \midrule
            Fully ft & 12.20 & 0.51 & 0.49 \\
            LoRA & 13.54 & 0.49 & 0.47 \\
            Merged & 11.69 & 0.50 & 0.49 \\
            Merged* & 11.17 & 0.51 & 0.49 \\
            \bottomrule
        \end{tabular}
    \end{minipage}\hfill
    \begin{minipage}{0.45\linewidth}
        \centering
        \includegraphics[width=0.8\linewidth]{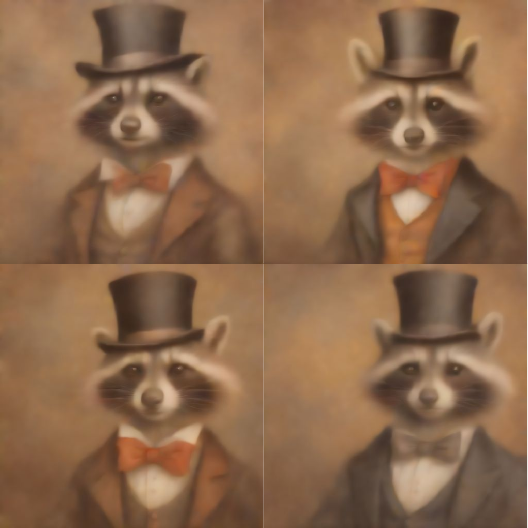}
        \caption{Mode collapse w/ SD Turbo training strategy.}
        \label{fig:rebuttal_modecollapse}
    \end{minipage}
\end{figure}

Our work aims to enhance the one-step diffusion models' performance. When no pretrained one-step model is available, we can still run the SwiftBrush (SB) training procedure on a small prompt dataset to build that initial model. To validate SBv2's effectiveness in that scenario, we re-train our method but using SB pretrained weights for initialization and report the results in \cref{tab:rebuttal}. As shown, our final merged model obtains the FID score of 11.69 without image regularization and 11.17 with image regularization. Note that DMD achieves FID of 11.49 with regularization on real images too. This result demonstrates that SBv2 still can reach SoTA one-step performance w/o the help of existing one-step models, though the gain is not as significant. Even when counting SB training time, our pipeline in this setting is still much more efficient than other one-step distillation methods.

\subsection{Compare to SD Turbo training with SB initialization.}
Three reasons make this training scheme less favorable than our approach: (1) SD Turbo training needs a large text-image pair dataset, which is costly in storage and computation, while SB training is image-free, making it more efficient and scalable. (2) SD Turbo's training code isn't publicly available, making reproduction difficult, whereas SB's training scheme much easier to reimplement and use. (3) SD Turbo's adversarial training is prone to mode collapse, requiring careful monitoring. Our attempts to reimplement SD Turbo faced mode collapse issues in early epochs (\cref{fig:rebuttal_modecollapse}).

\input{tables/supp_train}

\subsection{Training Cost and Inference Speed}
\myheading{Setup.} All of the self-measurements are taken on NVIDIA A100 40GB GPUs. However, most of the reported numbers about training time are taken directly from the corresponding papers, which did not specify whether A100 40GB or A100 80GB GPUs were used during training, except for SwiftBrush families. Even though both of these GPUs have equivalent computational speed, the 80GB version is capable of larger training batch size, which can drastically improve the training time. For inference time, we re-run every method except for GigaGAN due to its public model unavailability. We follow a standardized procedure to ensure fair comparisons for inference times. First, we warm up the model by running 5 inference passes. Then, we perform inference 50 times, repeating this process 10 times and taking the average of the results. The inference flow for distilled one-step diffusion models consists of three main steps: text encoding, UNet feedforwarding, and VAE decoding, without the denoising step since all models are one-step. To optimize memory usage and computational efficiency, all input data and model parameters are stored in float16 format throughout the inference process. 

\myheading{Analysis.} We sum up the training and inference time of the existing methods, including GAN-based, multi-step diffusion models, and one-step diffusion models in \cref{tab:traintime}. Notice that all of the one-step diffusion models are in some form of distillation from the teacher model, hence the training time gap. Among the distillation-based methods, SwiftBrush and our proposed approach stand out for their exceptional performance. SwiftBrush achieves an impressive inference time of 0.13s while maintaining a competitive FID score of 15.46. Our method, while maintaining the same inference speed, further improves the FID metrics, surpassing even the computationally intensive StyleGAN-T and GigaGAN models and the multi-step teacher. Although our method requires a slightly longer training time compared to SwiftBrush, the superior quality of the generated images, as evidenced by the significantly lower FID and other analyses in the main paper, justifies the additional training effort. These results demonstrate the effectiveness of our approach in achieving state-of-the-art image generation quality while maintaining near real-time inference speeds. Note that for methods using SDv1.5 as the teacher \cite{DMD,InstaFlow,ufo, HIPA}, the inference time gap mostly comes from the used text encoder, which is smaller than SDv2.1 based teacher.

\begin{figure}[h]
    \centering
    \includegraphics[width=\linewidth]{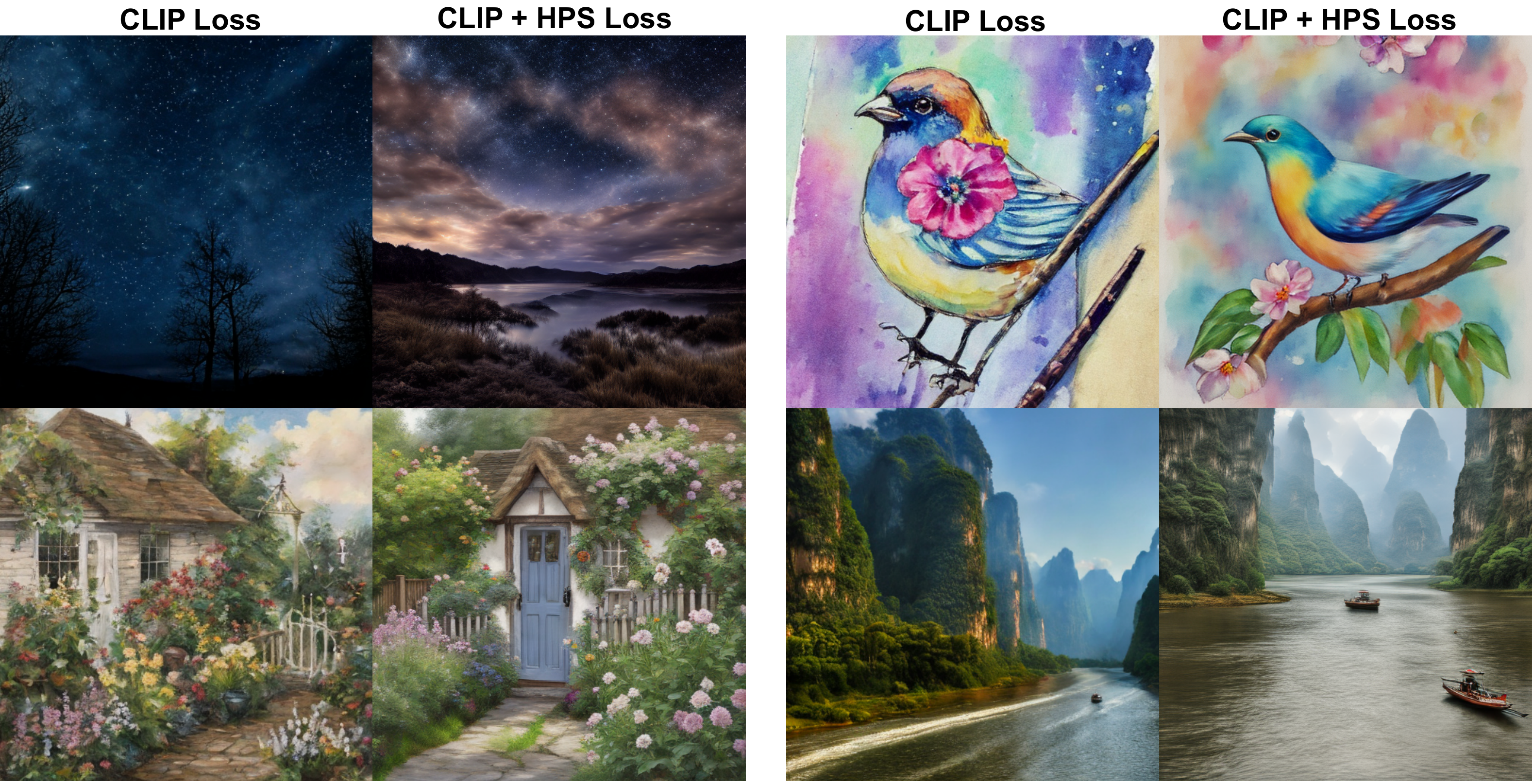}
    \caption{\textbf{Qualitative comparison} between samples training SwiftBrush with auxiliary losses: CLIP vs CLIP+HPS.}
    \label{fig:clipvshps}
\end{figure}
\input{tables/supp_hpscompare}

\section{Analysis and further applications}
\subsection{About the robustness of the training scheme}

We further demonstrate the robustness and flexibility of our distillation scheme, as stated in the main paper, by incorporating additional loss functions. 
Observing that the student can still improve upon the human preference aspect in the HPS benchmark, we integrate an HPSv2 loss, similar to the integration of the CLIP loss: We calculate the HPS score using decoded student predictions and apply a clamped ReLU loss with a weighting scheme for fine-grained control into the current resource-efficient distillation process (which already embeds the CLIP loss).

The results presented in \cref{tab:supp_hpscompare} show that by additionally integrating the HPSv2 loss, we successfully improve the model's performance on the HPSv2 benchmark under its zero-shot protocol while maintaining other metrics. This showcases the robustness of the SwiftBrush distillation process, enabling us to incorporate additional guidance to surpass the performance of the teacher model. \cref{fig:clipvshps} illustrates that by including HPS loss, the overall color of the generated images is more soothing for human eyes, enforcing the quantitative result improvement in the HPSv2 benchmark. However, upon closer examination of \cref{tab:supp_hpscompare}, we observe slight trade-offs in FID, suggesting the need for further investigation to enhance the synergy between these loss functions for it to surpass our current proposed solution. Therefore, we leave it as a direction for future research.

\subsection{Composition}
\label{sec:supp_composition}
\begin{figure}[t]
    \centering
    \includegraphics[width=1.\linewidth]{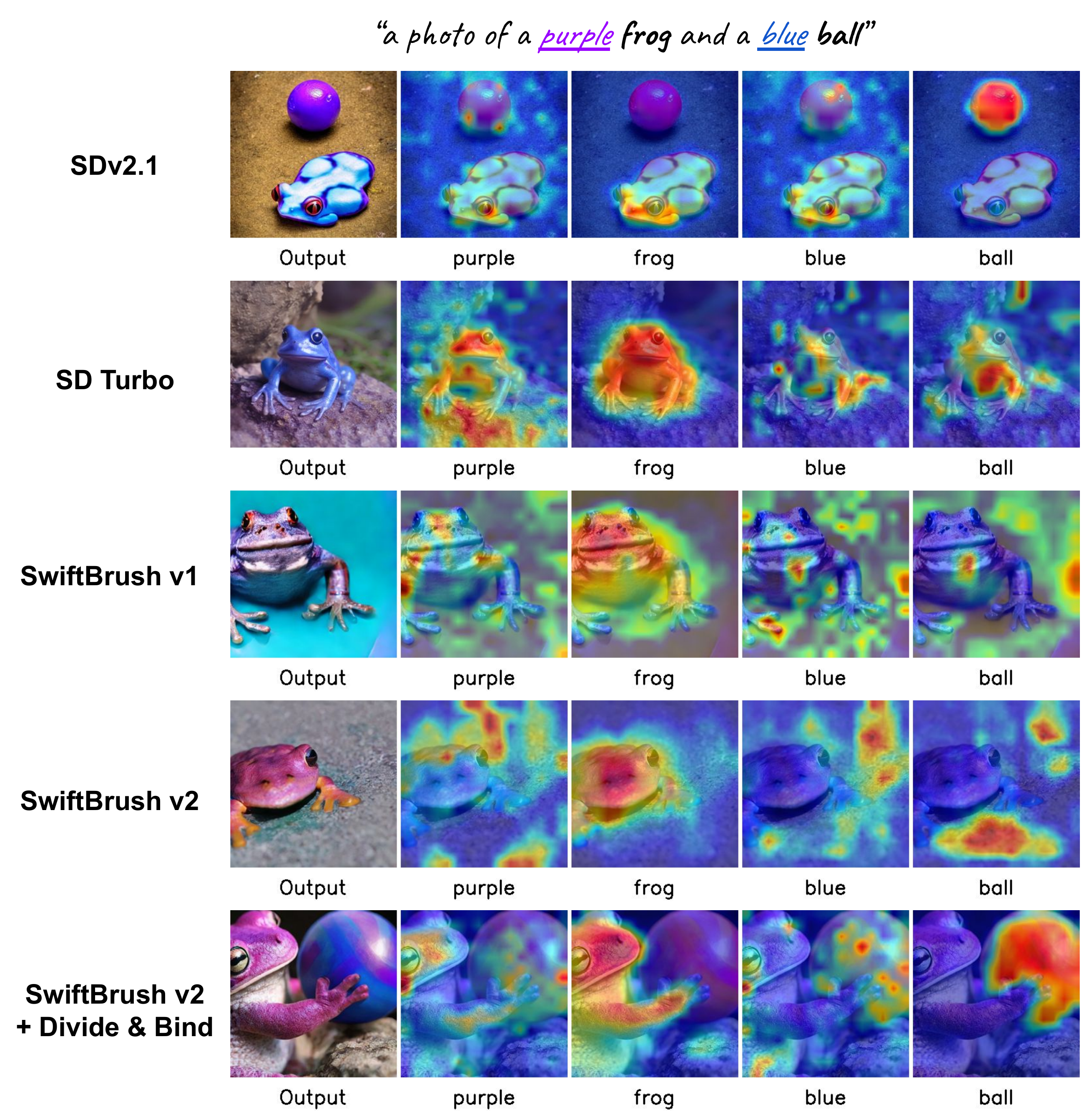}
    \caption{Cross-attention visualization for each object and attribute tokens between our models and the multi-step teacher SDv2.1, SD Turbo and SwiftBrush, generated with the same random seeds for the following prompt: ``A photo of a purple frog and a blue ball''.}
    \label{fig:badcomposecompare}
\end{figure}

In \cref{fig:badcomposecompare}, similar to other diffusion models, our distilled model still demonstrates low compositional ability and text-image misalignment when tasked with prompts that require generating multiple objects associated with attributes. Our model can generate the purple frog; however, it fails to generate the ball as well as its color. There have been a number of works aimed at solving this issue by running attention guidance \cite{controldiff, improvsq} or enhancing attention masks \cite{astar, excite, DivideB}. In our experiments, we chose to apply the Divide-and-Bind \cite{DivideB} approach since this method enhances both the generation of the objects and their corresponding attributes. Since the model predicts the final image in only one step, we choose to iteratively update the latent 100 times while keeping the same scale size of 20 as the original paper. As illustrated in the final row of \cref{fig:badcomposecompare}, our model demonstrates the ability to generate following the prompt accurately. The technique, initially developed for multi-step diffusion models, has been successfully adapted to enhance the output quality of our one-step model with minimal modifications. This achievement highlights the remarkable versatility and compatibility of our model with existing techniques commonly associated with text-to-image diffusion models. Although the original optimization process was designed for a multi-step approach, resulting in slower running times, our model demonstrates faster performance compared to the original multi-step teacher model. Moving forward, we believe that exploring novel latent optimization methods tailored explicitly for one-step diffusion models presents a promising new research direction, and we hope to draw attention to this area in the future.


\subsection{Latent Interpolation}
\begin{figure}[t]
    \small
    \begin{center}
        \includegraphics[width=.85\textwidth]{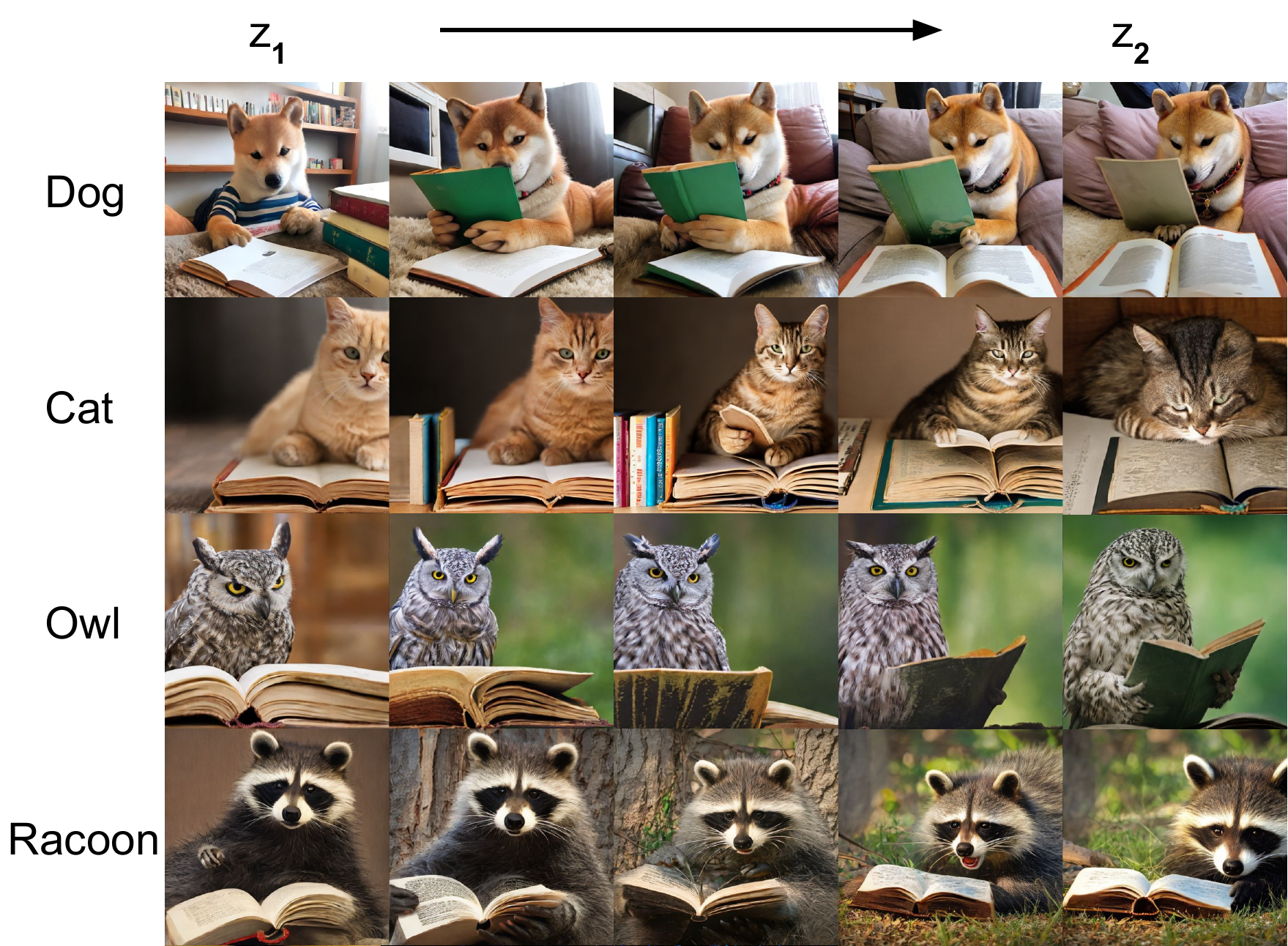}
    \end{center}
    \caption{Results of \textbf{interpolating the noise input}. The prompts
used here are selected from a standard template: \textit{``A photo of a \{animal\} reading a book''}. Here, ‘animal’ is dog, cat, owl or racoon and we interpolate the noise input using spherical linear interpolation (Slerp). Same text input $y$ is used for images at each row.}
    \label{fig:supp_interpolate_noise}
\end{figure}

By interpolating between two random input noises via spherical linear interpolation (slerp) in \cref{fig:supp_interpolate_noise}, our model seamlessly captures the gradual transformations in visual features without compromising adherence to the given prompt. This not only showcases the robustness of our model but also highlights the versatile image synthesis capabilities stemming from a well-trained conditional text-to-image model.

\subsection{Prompt Interpolation}
\begin{figure}[b]
    \small
    \begin{center}
        \includegraphics[width=.85\textwidth]{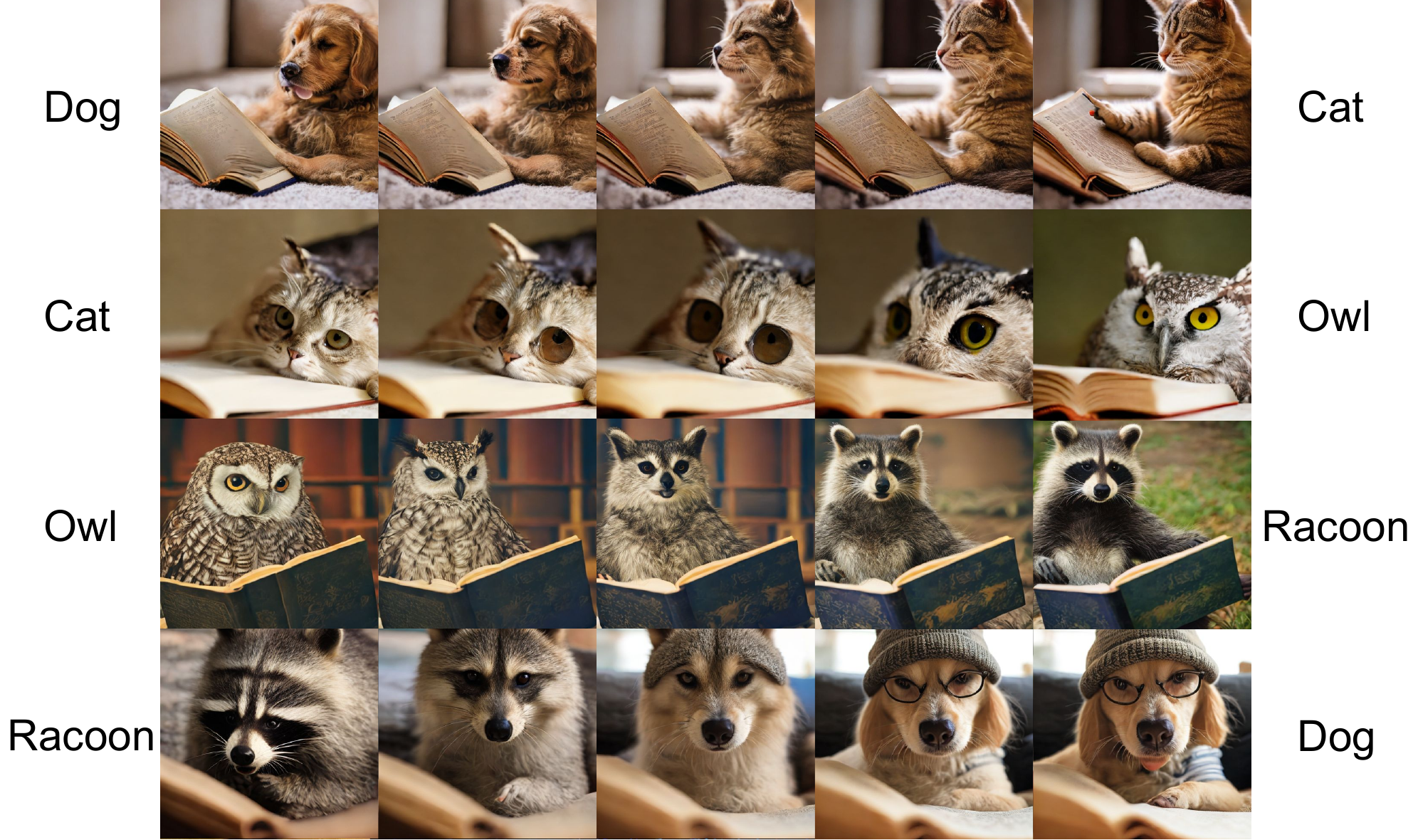}
    \end{center}
    \caption{Results of \textbf{interpolating the input prompt}. The prompts
used here are selected from a standard template: \textit{``A photo of a \{animal\} reading a book''}. Here, the text embedding is interpolated using spherical linear interpolation (Slerp). We interpolate two prompts using the same template but two different ‘animal’ (from left to right) and same noise input $z$ is used for images at each row.}
    \label{fig:supp_interpolate_text}
\end{figure}

We showcase our model's capability when interpolating two input prompt embeddings in  \cref{fig:supp_interpolate_text} using same template but with only one word different. Our model effectively captures and blends the characteristics of both prompts, creating visually compelling and coherent intermediate representations. This feature demonstrates our model's capacity to understand and manipulate the semantic relationships between different textual inputs, providing a powerful tool for creative exploration and generating diverse images that seamlessly bridge different concepts or styles.

\subsection{Arbitrary size and aspect ratio generation with ScaleCrafter}
\begin{figure}[t]
    \centering
    \includegraphics[width=\textwidth]{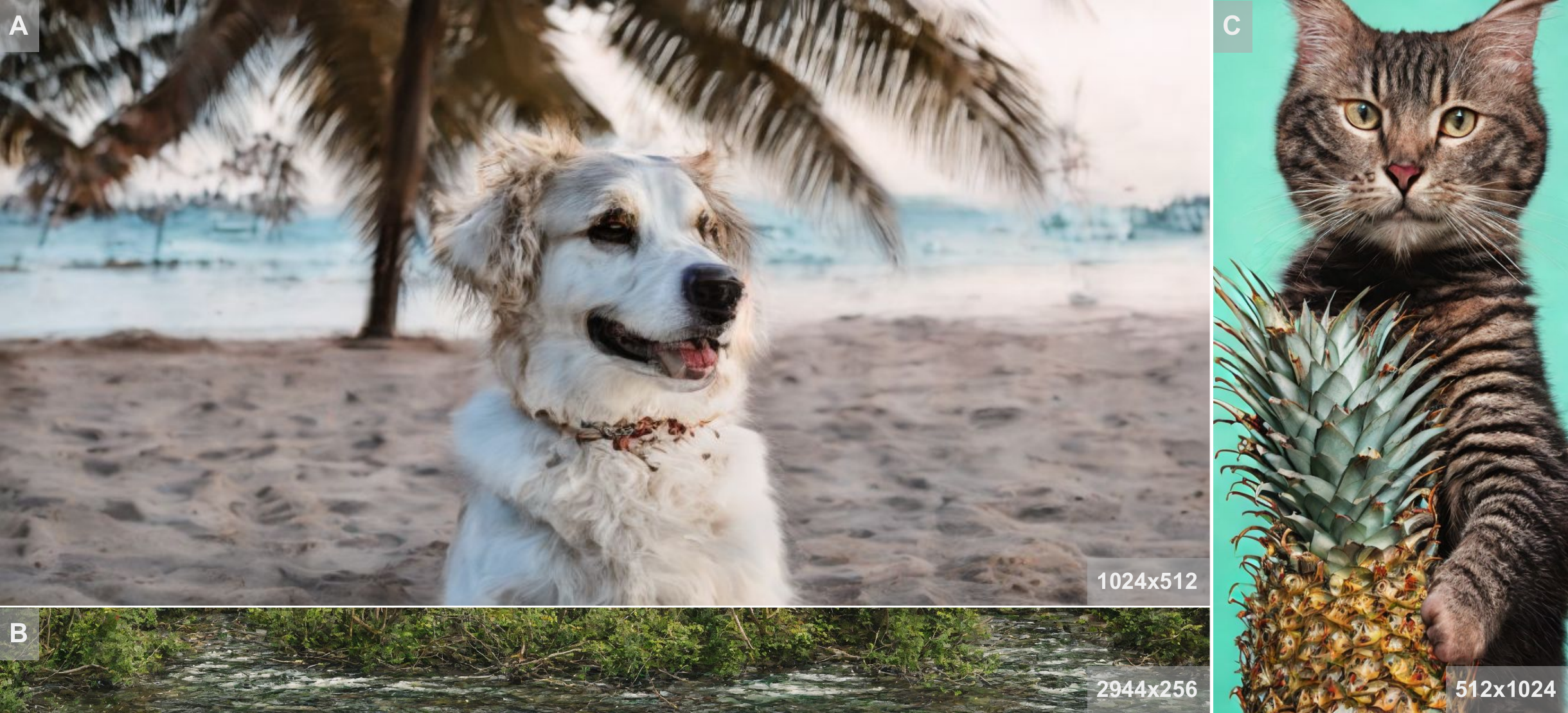}
    \caption{Qualitative results of our model combined with ScaleCrafter. Here, the caption for \textbf{A} is ``A dog sitting on a beautiful beach, with palm trees behind, bokeh'', that for \textbf{B} is ``A river running through a forest'' and that for \textbf{C} is ``Photo of a cat holding a pineapple''.}
    \label{fig:sup_scalecrafter_sbv2}
\end{figure}

Our models inherit the architecture of SDv2.1, which limits the ability to generate images with varying sizes or aspect ratios, unlike other works such as SD-XL \cite{sdxl,lin2024sdxllightning}. To address this limitation, we apply ScaleCrafter \cite{he2023scalecrafter}, a technique that adjusts the convolution layers of the U-Net model during inference through re-dilation. \cref{fig:sup_scalecrafter_sbv2} illustrates the synthesized images generated in various ratios and resolutions using this method. This once again demonstrates that our one-step model can effectively incorporate techniques from the multi-step diffusion model domain to enhance performance, similar to the application of Divide-and-Bind for improving composition, as discussed in \cref{sec:supp_composition}.

\begin{figure}[t]
    \centering
    \includegraphics[width=\textwidth]{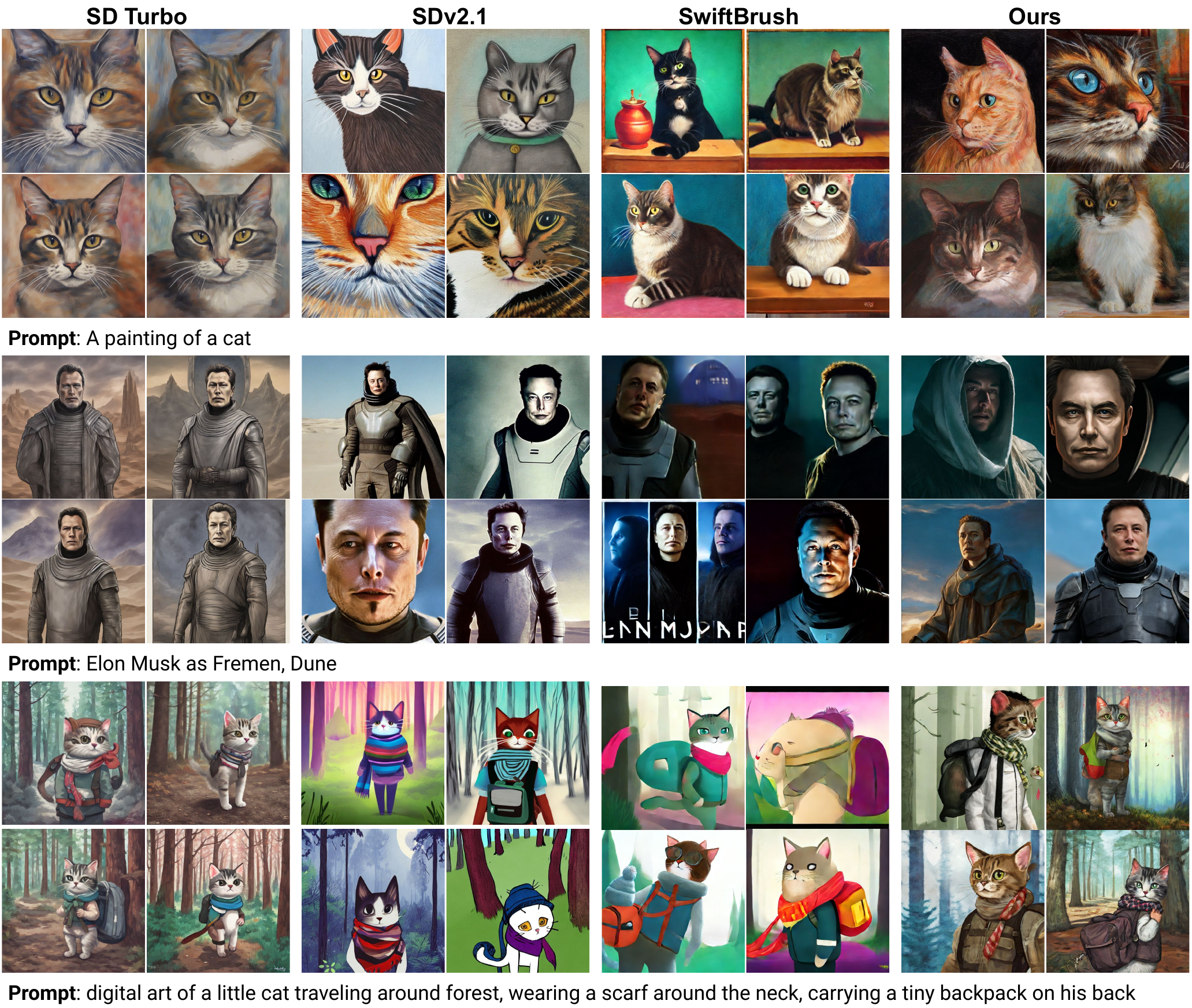}
    \caption{\textbf{Qualitative comparison} between our models and the multi-step teacher SDv2.1, SD Turbo and SwiftBrush. }
    \label{fig:sup_qual_compare}
\end{figure}


\section{More qualitative results}

\begin{figure}[t]
    \centering
    \includegraphics[width=0.9\textwidth]{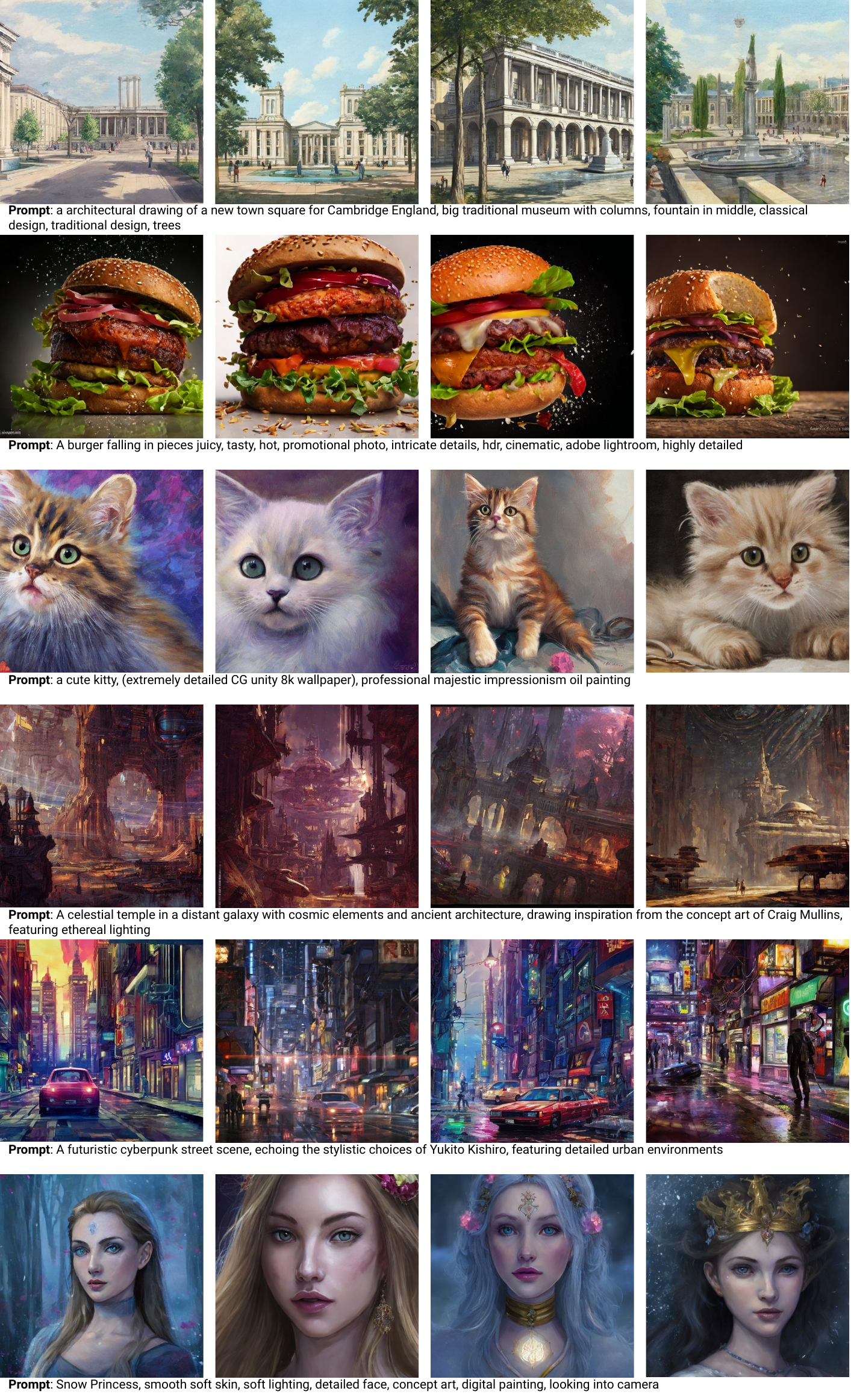}
    \caption{\textbf{Diversity.} Our models generate realistic, diverse images spanning various object categories, styles, and scenes.}
    \label{fig:sup_diversity_sbv2}
\end{figure}

\begin{figure}[t]
    \centering
    \includegraphics[width=\textwidth]{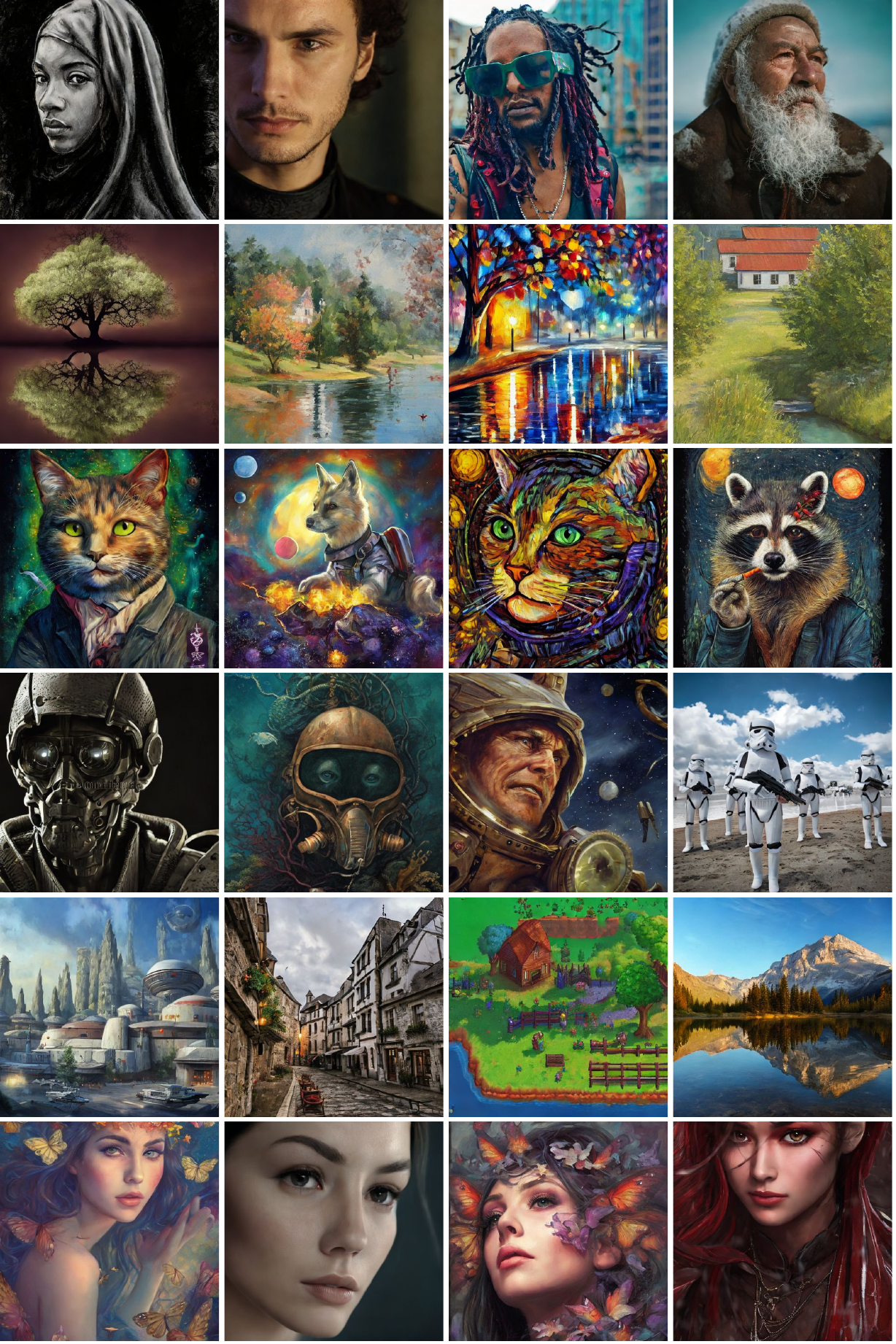}
    \caption{ Additional qualitative images generated by our model.}
    \label{fig:sup_spam}
\end{figure}

We provide additional comparisons of our model with SD Turbo, SDv2.1, and SwiftBrush in \cref{fig:sup_qual_compare}. \cref{fig:sup_diversity_sbv2} illustrates the diversity of our model when generating images with the same input prompts. Finally, \cref{fig:sup_spam} shows more uncurated samples synthesized by our model.

\end{document}

%% file: tables/datasetcompare.tex
\begin{table}[t]
\centering
\setlength{\tabcolsep}{4pt}
\caption{\textbf{The impact of SD Turbo initialization and dataset size} on SwiftBrush's performance, compared to SD Turbo on the zero-shot COCO-2014 benchmark.}
\label{tab:datasetcompare}
\begin{tabular}{lccccc}
\toprule
\textbf{Model Name} & \textbf{Training size} & \textbf{FID}$\downarrow$ & \textbf{CLIP}$\uparrow$ & \textbf{Precison}$\uparrow$ & \textbf{Recall}$\uparrow$ \\ 
\midrule
SD Turbo   & -  & 16.10             & \textbf{0.33}    & \textbf{0.65}    & 0.35             \\
SwiftBrush & 1.3M & 15.46 & 0.30             & 0.47             & 0.46    \\
SwiftBrush (Turbo init.) & 1.3M & \underline{14.59} & 0.30             & 0.43             & \textbf{0.55}    \\
SwiftBrush (Turbo init.) & 3.3M & \textbf{11.27}    & \underline{0.31} & \underline{0.48} & \underline{0.54} \\
\bottomrule
\end{tabular}

\end{table}

%% file: tables/work_pr_compare.tex
\begin{table}[t]
\centering
\footnotesize
\setlength{\tabcolsep}{7pt}
\caption{\textbf{Quantitative comparisons} between our method and others on zero-shot MS COCO-2014 benchmark. For multi-step SD models, we report each with the \textit{cfg} that returns the best FID, e.g., \textit{cfg} = 3 for SDv1.5 and \textit{cfg} = 2 for SDv2.1. We also report performance of the teacher model (SDv2.1 with \textit{cfg} = 4.5). $^\dag$ denotes reported numbers, $^\ddag$ denotes our rerun based on provided GitHubs. `-' denotes unreported data. Ours* indicates training with additional image regularization.
}
\begin{tabular}{lccccc}
\toprule
\multicolumn{1}{c}{\textbf{Method}} & 
\textbf{NFEs} & 
\textbf{FID$\downarrow$} & \textbf{CLIP$\uparrow$} & \textbf{Precision$\uparrow$} & \textbf{Recall$\uparrow$} \\ 
\midrule
StyleGAN-T \cite{sauer2023stylegan}$^\dag$ & 1 & 13.90            & -              & -                & -             \\
GigaGAN \cite{kang2023scaling}$^\dag$      & 1 & 9.09             & -              & -                & -             \\ 
\midrule
SDv1.5 \cite{SD} (\textit{cfg} = 3)$^\dag$                    & 25 & 8.78             & 0.30              & 0.59             & \textbf{0.53} \\
SDv2.1 \cite{SD} (\textit{cfg} = 2)$^\ddag$                   & 25 & 9.64            & 0.31 & 0.57 & \textbf{0.53}          \\
SDv2.1 \cite{SD} (\textit{cfg} = 4.5)$^\ddag$                   & 25 & 12.26            & \textbf{0.33} & \underline{0.61} & 0.41          \\
\midrule
SD Turbo \cite{SDTurbo}$^\ddag$            & 1 & 16.10            & \textbf{0.33}    & \textbf{0.65}    & 0.35          \\ 
UFOGen \cite{ufo}$^\dag$                   & 1 & 12.78            & -              & -                & -             \\
MD-UFOGen \cite{mobilediffusion}$^\dag$    & 1 & 11.67            & -              & -                & -             \\
HiPA \cite{HIPA}$^\dag$                    & 1 & 13.91            & 0.31             & -                & -             \\
InstaFlow-0.9B \cite{InstaFlow}$^\ddag$    & 1 & 13.33            & 0.30             & 0.53             & 0.45          \\
DMD \cite{DMD}$^\dag$                                 & 1 & 11.49            & \underline{0.32} & -                & -             \\
SwiftBrush                               & 1 & 15.46            & 0.30             & 0.47             & 0.46          \\ 
\midrule
Ours                                     & 1 & \underline{8.77} & \underline{0.32} & 0.55             & \textbf{0.53} \\
Ours*                     & 1 & \textbf{8.14}    & \underline{0.32} & 0.57             & \underline{0.52}          \\ \bottomrule
\end{tabular}
\label{tab:coco_compare}
\end{table}

%% file: tables/hps.tex
\begin{table}[t]
\centering
\footnotesize
\setlength{\tabcolsep}{5pt}
\caption{\textbf{HPSv2 comparisons} between our method and previous work. $^{\dag}$ denotes reported numbers, $^{\ddag}$ denotes our rerun based on provided GitHubs. Ours* indicates training with additional image regularization. \textbf{Bold} indicates the best, while \underline{underline} indicates the second best.}
\begin{tabular}{lcccc}
\toprule
\textbf{Method} & \textbf{Anime}    & \textbf{Photo}    & \textbf{Concept Art} & \textbf{Painting} \\ 
\midrule
SDv2.1$\dag$    & \underline{27.48} & 26.89             & \underline{26.86}    & \textbf{27.46}    \\ 
\midrule
SD Turbo$\ddag$             & \textbf{27.98}    & \underline{27.59} & \textbf{27.16}       & \underline{27.19} \\
InstaFlow$\dag$ & 25.98 & 26.32          & 25.79             & 25.93 \\
BOOT$\dag$      & 25.29 & 25.16          & 24.40             & 24.61 \\
SwiftBrush$\dag$         & 26.91 & 27.21          & 26.32             & 26.37 \\ 
\midrule
Ours            & 27.13 & 27.56          & 26.69             & 26.76 \\
Ours*           & 27.25 & \textbf{27.62} & \underline{26.86} & 26.77 \\
\bottomrule
\end{tabular}
\label{tab:hps_compare}
\end{table}

%% file: tables/config.tex
\begin{table}[t]
\centering
\footnotesize
\setlength{\tabcolsep}{3pt}
\caption{\textbf{Ablation of our methods} upon zero-shot MS COCO-2014 30K. \textbf{Bold} indicates the best, while \underline{underline} indicates the second best.}
\label{tab:ablation}
\resizebox{.95\textwidth}{!}{
\begin{tabular}{lccccccc}
\toprule
\textbf{Label} &
  \textbf{SD Turbo} &
  \textbf{CLIP} &
  \textbf{LAION} &
  \textbf{FID}$\downarrow$ &
  \textbf{CLIP}$\uparrow$ &
  \textbf{Precision}$\uparrow$ &
  \textbf{Recall}$\uparrow$ \\ 
  \midrule
  SwiftBrush &  &   &             & 15.46                & 0.30                & 0.47                 & 0.46                 \\ 
\midrule
\rowcolor{gray!30} \multicolumn{8}{c}{Fully training}         \\
\textbf{}  & \cmark             &                  &                  & 14.59                & 0.29                & 0.43                 & \textbf{0.55}        \\
A  & \cmark             &                  & \cmark           & 11.27                & 0.31                & 0.48                 & \underline{0.54}     \\ 
\midrule
 \rowcolor{gray!30} \multicolumn{8}{c}{Efficient training}  \\
\textbf{}  & \cmark             &            &                  & 13.21                & 0.32       & 0.61        & 0.38                 \\
\textbf{}  & \cmark             & \cmark           &                  & 11.70                & \textbf{0.33}       & \textbf{0.63}        & 0.42                 \\
B  & \cmark             & \cmark           & \cmark           & 11.02                & 0.32                & 0.51                 & 0.52                 \\
\midrule
Ours & \multicolumn{3}{l}{Merge A and B}                & \underline{8.77}     & 0.32                & 0.55                 & 0.53                 \\
Ours* & \multicolumn{3}{l}{Merge A and B w/ regularization} & \textbf{8.14}        & \underline{0.32}    & \underline{0.57}     & 0.52   \\
\bottomrule
\end{tabular}
}
\label{tab:ablate_all}
\end{table}

%% file: tables/clipcompare.tex

\begin{table}[t]
\centering
\footnotesize
\setlength{\tabcolsep}{7pt}
\caption{\textbf{Ablation} of our enhanced CLIP loss and resource-efficient training scheme.}
\resizebox{.95\textwidth}{!}{
\begin{tabular}{cccccccc}
\toprule
\textbf{CLIP} & \textbf{Clamped} & \textbf{Scheduler} & 
  \textbf{Efficient} & 
  \textbf{FID$\downarrow$} &
  \textbf{CLIP$\uparrow$} &
  \textbf{GPU days} \\ 
\midrule
 & & & &14.59          & 0.297             & \textbf{4.1}    \\
\midrule
\cmark   &          &           &        & 21.32          & \textbf{0.337}    & 7.8             \\
\cmark   & \cmark   &          &         & 13.19          & 0.319             & 7.8             \\
\cmark   & \cmark   & \cmark & \cmark   & \textbf{11.70} & 0.330             & \underline{4.3} \\
\bottomrule
\end{tabular}
}
\label{tab:clipcompare}
\end{table}

%% file: tables/supp_train.tex
\begin{table}[t]
\centering
\setlength{\tabcolsep}{5pt}
\caption{\textbf{Comparison of inference and training time} between our method against other works upon the zero-shot benchmark on MS COCO-2014. $^{\dag}$means that we obtain the numbers from the reports. $^{\ddag}$means that we obtain the numbers by ourselves.  The inference time in float16 precision for all methods was reproduced using a single NVIDIA A100 40GB to ensure a fair comparison. The units for training time were also calculated using NVIDIA A100. Model A is the fully-finetuned SwiftBrush longer and with more data (as described in the main paper), whereas Model B is trained with auxiliary loss by utilizing the resources-efficient training scheme. Note that in the case of HiPA, the model is trained with COCO-2017, hence not a zero-shot result; and for methods using SDv1.5 as the teacher \cite{DMD,InstaFlow,ufo, HIPA}, the inference time gap with ours mostly comes from the used text encoder, which is smaller than SDv2.1 based teacher.}
\begin{tabular}{lccccc}
\toprule
\multicolumn{1}{c}{\textbf{Method}} &
  \textbf{NFE} &
  \multicolumn{1}{l}{\textbf{Distill}} &
  \textbf{\begin{tabular}[c]{@{}c@{}}Inference \\ Time (s) $\downarrow$\end{tabular}} &
  \textbf{\begin{tabular}[c]{@{}c@{}}Training Time\\ (GPU days) $\downarrow$\end{tabular}} &
  \textbf{FID$\downarrow$} \\ \hline
StyleGAN-T$^{\dag}$\cite{sauer2023stylegan}                & 1  & \xmark & 0.10  & 1792.0  & 13.90 \\
GigaGAN$^{\dag}$\cite{kang2023gigagan}                   & 1  & \xmark & 0.13  & 6250.0  & 9.09  \\ \hline
SDv1.5$^{\dag}$\cite{SD}                   & 25 & \xmark & 1.74 & 4783.0  & 8.78  \\
SDv2.1(\textit{cfg}=4.5)$^{\dag}$\cite{SD} & 25 & \xmark & 1.77 & -     & 12.26 \\ \hline
InstaFlow-0.9B$^{\dag}$\cite{InstaFlow}            & 1  & \cmark & 0.12   & 108.0   & 13.10 \\
UFOGen$^{\dag}$\cite{ufo}                    & 1  & \cmark & 0.12   & -     & 12.78 \\
DMD$^{\dag}$\cite{DMD}                       & 1  & \cmark & 0.12   & 108.0 & 11.49 \\
HiPA$^{\dag}$\cite{HIPA}                      & 1  & \cmark & -    & 3.8   & 13.91 \\
SD Turbo$^{\dag}$\cite{SDTurbo}                      & 1  & \cmark & 0.13    & -   & 16.10 \\
SwiftBrush$^{\ddag}$\cite{SwiftBrush}                      & 1  & \cmark & 0.13   & 4.1   & 15.46 \\ \hline
Model A$^{\ddag}$                         & 1  & \cmark & 0.13   & 12.1  & 11.27 \\
Model B$^{\ddag}$                         & 1  & \cmark & 0.13   & 12.0  & 11.02 \\
Ours (A+B)$^{\ddag}$                            & 1  & \cmark & 0.13   & 24.1  & 8.77 \\
Ours w/ regularizer$^{\ddag}$                            & 1  & \cmark & 0.13   & 24.1  & 8.14 \\
\bottomrule
\end{tabular}
\label{tab:traintime}
\end{table}


%% file: tables/supp_hpscompare.tex
\begin{table}[t]
\centering
\caption{\textbf{Comparison of metrics} between teacher, our method and our method with auxiliary loss HPS upon the zero-shot benchmark on MS COCO-2014. \textbf{Bold} means the best result while \underline{underline} means the second-best result.}
\begin{tabular}{lcccccc}
\toprule
\multirow{2}{*}{\textbf{Method}} &
  \multicolumn{1}{c}{\multirow{2}{*}{\textbf{FID}$\downarrow$}} &
  \multicolumn{1}{c}{\multirow{2}{*}{\textbf{CLIP}$\uparrow$}} &
  \multicolumn{4}{c}{\textbf{Human Preference Score v2}$\uparrow$} \\ \cmidrule{4-7} 
 &
  \multicolumn{1}{c}{} &
  \multicolumn{1}{c}{} &
  \multicolumn{1}{c}{\textbf{Anime}} &
  \multicolumn{1}{c}{\textbf{Photo}} &
  \multicolumn{1}{c}{\textbf{Concept Art}} &
  \multicolumn{1}{c}{\textbf{Painting}} \\ \midrule
SDv2.1(\textit{cfg}=4.5) & 12.26             & 0.32 & \underline{27.55} & \underline{27.80} & \underline{26.85} & \underline{26.73} \\ 
Model B                  & \textbf{11.02}    & 0.32 & 26.55             & 26.71             & 26.17             & 26.11             \\
Model B + HPSv2          & \underline{11.15} & 0.32 & \textbf{28.22}    & \textbf{28.00}    & \textbf{27.35}    & \textbf{27.42}   \\
\bottomrule
\end{tabular}
\label{tab:supp_hpscompare}
\end{table}